%% file: googledeepmind-test.tex
\documentclass[11pt, a4paper, logo, twocolumn, copyright,]{googledeepmind}

\pdftrailerid{redacted}

\makeatletter
\renewcommand\bibentry[1]{\nocite{#1}{\frenchspacing\@nameuse{BR@r@#1\@extra@b@citeb}}}
\makeatother

\usepackage{kantlipsum, lipsum}
\usepackage{dsfont}
\usepackage{gdm-colors}

\usepackage[authoryear, sort&compress, round]{natbib}

\graphicspath{{figures/}}

\title{Avoiding spurious sharpness minimization broadens applicability of \SAM}

\correspondingauthor{contact@sidakpal.com, hmobahi@google.com, thetish@google.com, ynd@google.com}


\usepackage{microtype}
\usepackage{graphicx}
\usepackage{subfigure}
\usepackage{booktabs} 

\usepackage{hyperref}

\usepackage{amsmath}
\usepackage{amssymb}
\usepackage{mathtools}
\usepackage{amsthm}

\usepackage{multirow}
\usepackage[capitalize,noabbrev]{cleveref}

\theoremstyle{plain}

\theoremstyle{definition}

\theoremstyle{remark}

\usepackage[textsize=tiny]{todonotes}

\input{dimensions}

\input{math_defs}

\usepackage{makecell}

\newcommand{\rad}{\rho} 
\renewcommand{\th}{\m{\theta}}

\newcommand{\eps}{\m{\epsilon}}

\newcommand{\logitText}{\textbf{\texttt{logit}}\xspace}
\newcommand{\funcText}{\textbf{\texttt{func}}\xspace}
\newcommand{\crossText}{\textbf{\texttt{cross}}\xspace}
\newcommand{\glog}{\mathbf{\delta}_{\logitText}} 
\newcommand{\gfun}{\mathbf{\delta}_{\funcText}}
\newcommand{\tlog}{\tau_{\logitText}}
\newcommand{\tfn}{\tau_{\funcText}}
\newcommand{\tc}{\tau_{\crossText}}
\newcommand{\SP}{\textbf{\texttt{SP}}\xspace}

\usepackage{listings}
\definecolor{codegreen}{rgb}{0,0.6,0}
\definecolor{codegray}{rgb}{0.5,0.5,0.5}
\definecolor{codepurple}{rgb}{0.58,0,0.82}
\definecolor{backcolour}{rgb}{0.95,0.95,0.92}

\lstdefinestyle{mystyle}{
  backgroundcolor=\color{backcolour}, commentstyle=\color{codegreen},
  keywordstyle=\color{magenta},
  numberstyle=\tiny\color{codegray},
  stringstyle=\color{codepurple},
  basicstyle=\ttfamily\footnotesize,
  breakatwhitespace=false,         
  breaklines=true,                 
  captionpos=b,                    
  keepspaces=true,                 
  showspaces=false,                
  showstringspaces=false,
  showtabs=false,                  
  tabsize=2
}

\newif\ifcomments
\commentsfalse 

\ifcomments
\newcommand{\ynd}[1]{{\color{blue}[YD: #1]}}
\newcommand{\hmb}[1]{{\color{purple}[HM: #1]}}
\newcommand{\aga}[1]{{\color{red}[AA: #1]}}
\newcommand{\sps}[1]{{\color{olive}[SPS: #1]}}
\else
\newcommand{\ynd}[1]{}
\newcommand{\hmb}[1]{}
\newcommand{\aga}[1]{}
\newcommand{\sps}[1]{}
\fi

\usepackage[dvipsnames]{xcolor}

\usepackage[most]{tcolorbox}
\usepackage{fontawesome} 
\usepackage[]{enumitem}

\reportnumber{} 

\author[*,1]{Sidak Pal Singh}
\author[2]{Hossein Mobahi}
\author[2]{Atish Agarwala}
\author[2]{Yann Dauphin}

\affil[*]{Work done during internship at Google DeepMind}
\affil[1]{ETH Zürich}
\affil[2]{Google DeepMind}

\begin{abstract}
Curvature regularization techniques like Sharpness Aware Minimization (\SAM) have shown great promise in improving generalization on vision tasks. However, we find that \SAM performs poorly in domains like natural language processing (NLP), often degrading performance --- even with twice the compute budget. We investigate the discrepancy across domains and find that \textit{in the NLP setting, \SAM is dominated by regularization of the logit statistics --— instead of improving the geometry of the function itself}. We use this observation to develop an alternative algorithm we call \funcSAM , which regularizes curvature only through modification of the statistics of the overall function implemented by the neural network, and avoids spurious minimization through logit manipulation. Furthermore, we argue that preconditioning the \SAM perturbation also prevents spurious minimization, and when combined with \funcSAM, it gives further improvements. Our proposed algorithms show improved performance over \adamw and \SAM baselines when trained for an equal number of steps, in both fixed-length and Chinchilla-style training settings, at various model scales (including \textit{billion-parameter scale}). On the whole, our work highlights the importance of more precise characterizations of sharpness in broadening the applicability of curvature regularization to large language models (LLMs).\looseness=-1\vspace{-0.5em}
\end{abstract}

\begin{document}

\maketitle

\input{template_content}

\bibliographystyle{abbrvnat}
\nobibliography*
\bibliography{template_refs}

\appendix
\onecolumn

\section*{Supplementary Materials}
\section{Additional Results}

\label{app:sharp_plots}

\begin{figure}[ht!]
\centering
\begin{tabular}{cc}
\includegraphics[height=0.33\linewidth]{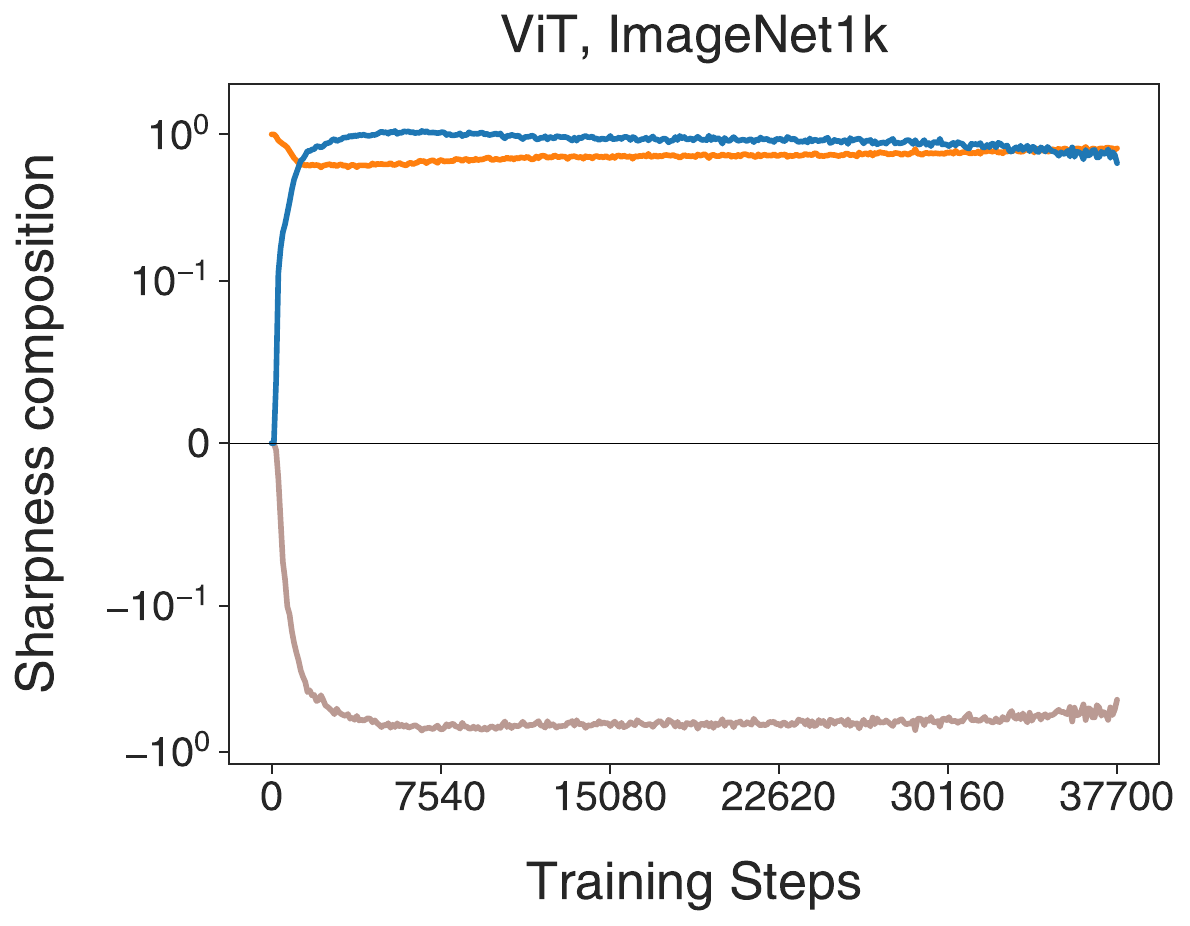} & \includegraphics[height=0.33\linewidth]{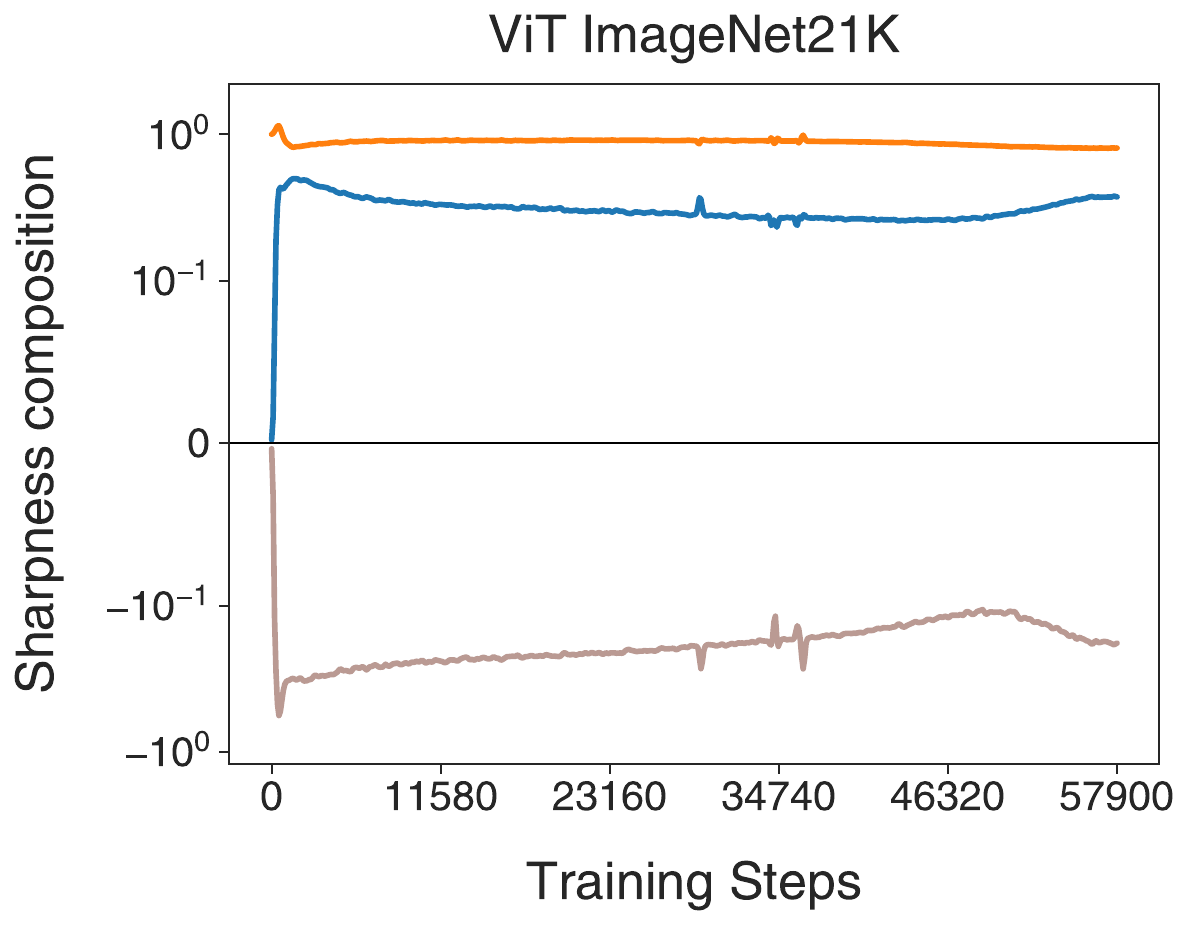} \\
\includegraphics[height=0.33\linewidth]{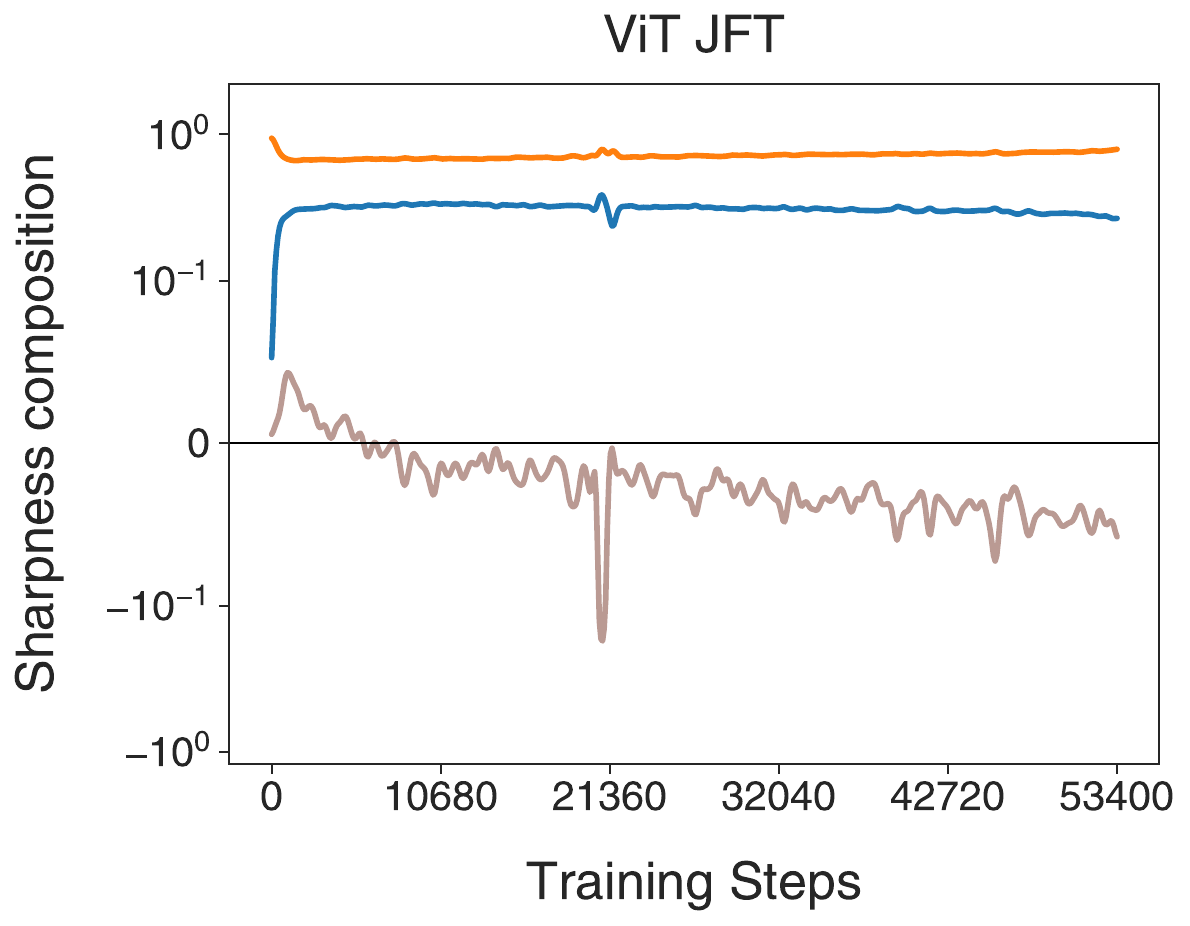} &
\includegraphics[height=0.33\linewidth]{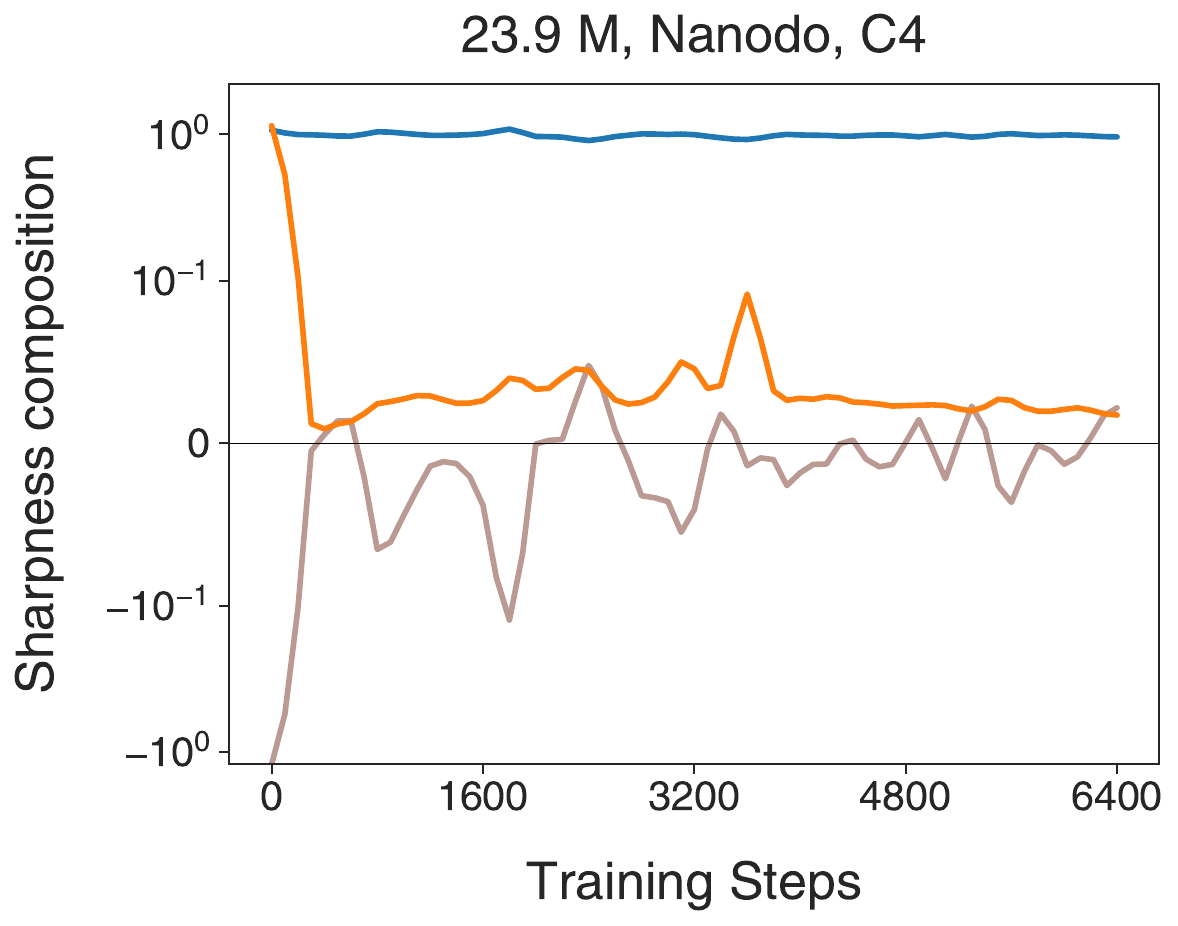} \\
\includegraphics[height=0.33\linewidth]{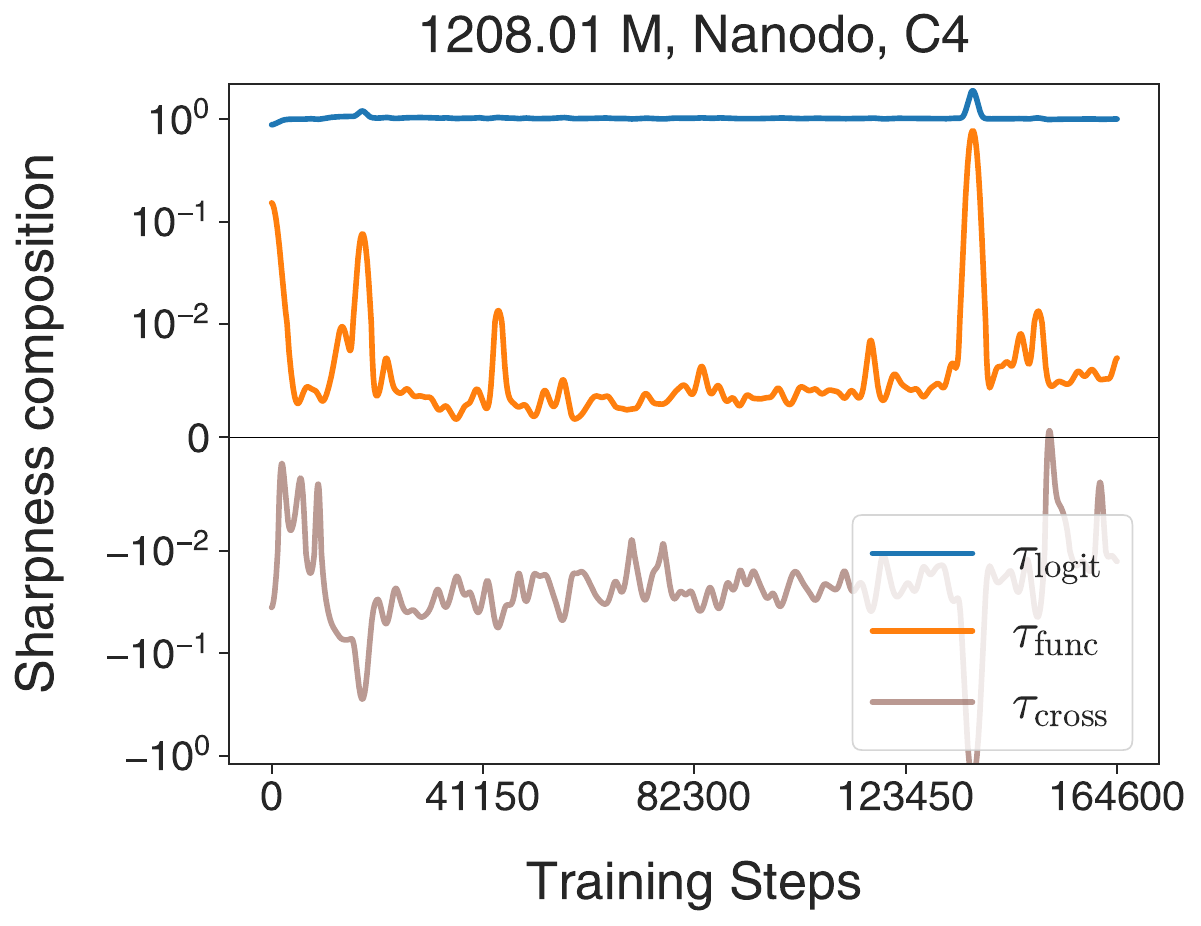}   
\end{tabular}
\caption{Sharpness contributions $\lcolor{\tlog}$, $\fcolor{\tfn}$ and $\tc$ for various datasets. $\tc$ tends to be negative for most of training.\looseness=-1}
\label{fig:sam_contribution_app}
\end{figure}


\begin{table}[ht!]
\caption{Comparison of \SAM variants across non-linearities for a Nanodo model with 2M (non-embedding parameters) on C4 dataset.}
\label{tab:non-linearities}
\centering
\begin{tabular}{@{}llc@{}}
\toprule
\textsc{Non-Linearity} & \textsc{Method}                                                        & \textsc{Eval Loss} \\ \midrule
GeLU          & \begin{tabular}[c]{@{}l@{}}\precond\\\funcSAM \end{tabular}                                                & 3.8614    \\[4mm]
GeLU          & \begin{tabular}[c]{@{}l@{}}\precond\\\SAM \end{tabular} & 3.8894    \\[4mm]
GeLU          & \adamw                                                         & 3.9069    \\[2mm] \midrule
ReLU          & \begin{tabular}[c]{@{}l@{}}\precond\\\funcSAM\end{tabular}                                                & 3.8777    \\[4mm]
ReLU          & \begin{tabular}[c]{@{}l@{}}\precond\\\SAM\end{tabular}  & 3.8937    \\[4mm]
ReLU          & \adamw                                                         & 3.9145    \\ \bottomrule
\end{tabular}
\end{table}

\begin{table}[h]
    \centering
    \caption{Comparison of different methods based on Hessian ${\HL}$ and GGN ${\HG}$ metrics for the $117.9$M model trained as per Chinchilla like training setup.}
    \renewcommand{\arraystretch}{1.2} 
\setlength{\tabcolsep}{4pt} 
    \begin{tabular}{lcccc}
        \toprule
        \textsc{Method} & \textsc{Eval Loss} & $\lambda_{\text{max}}({\HL})$ & $\mathrm{tr}({\HL})$ & $\mathrm{tr}({\HG})$ \\
        \midrule
        \makecell[l]{\precond\\ \funcSAMbrief} & \textbf{3.096} & 8.884 & \textit{2790.650} & 2746.559 \\[4mm]
        \makecell[l]{\precond \\ \SAM}  & 3.108 & \textit{7.415} & 2920.445 & \textit{2060.214} \\[4mm]
        \SAM& 3.126 & \textbf{3.381} & \textbf{2235.759} & \textbf{2222.779} \\[2mm]
        \adamw& 3.120 & 9.259 & 3299.497 & 3285.881 \\
        \bottomrule
    \end{tabular}
    \label{tab:hessian_comparison_117.9M}
\end{table}

\begin{figure*}[h]
\centering
\begin{tabular}{ccc}
\includegraphics[height=0.25\linewidth]{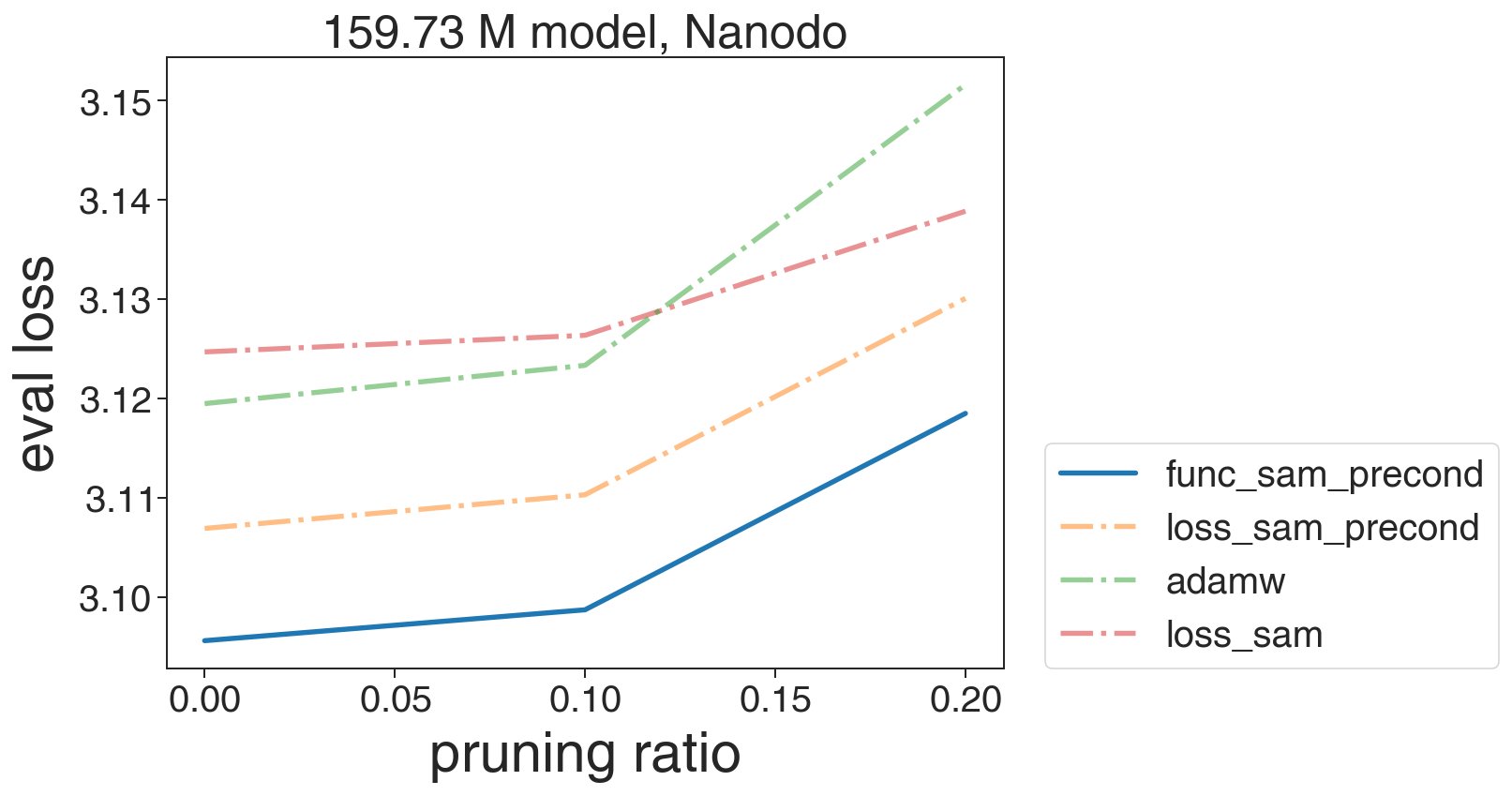} & \includegraphics[height=0.25\linewidth]{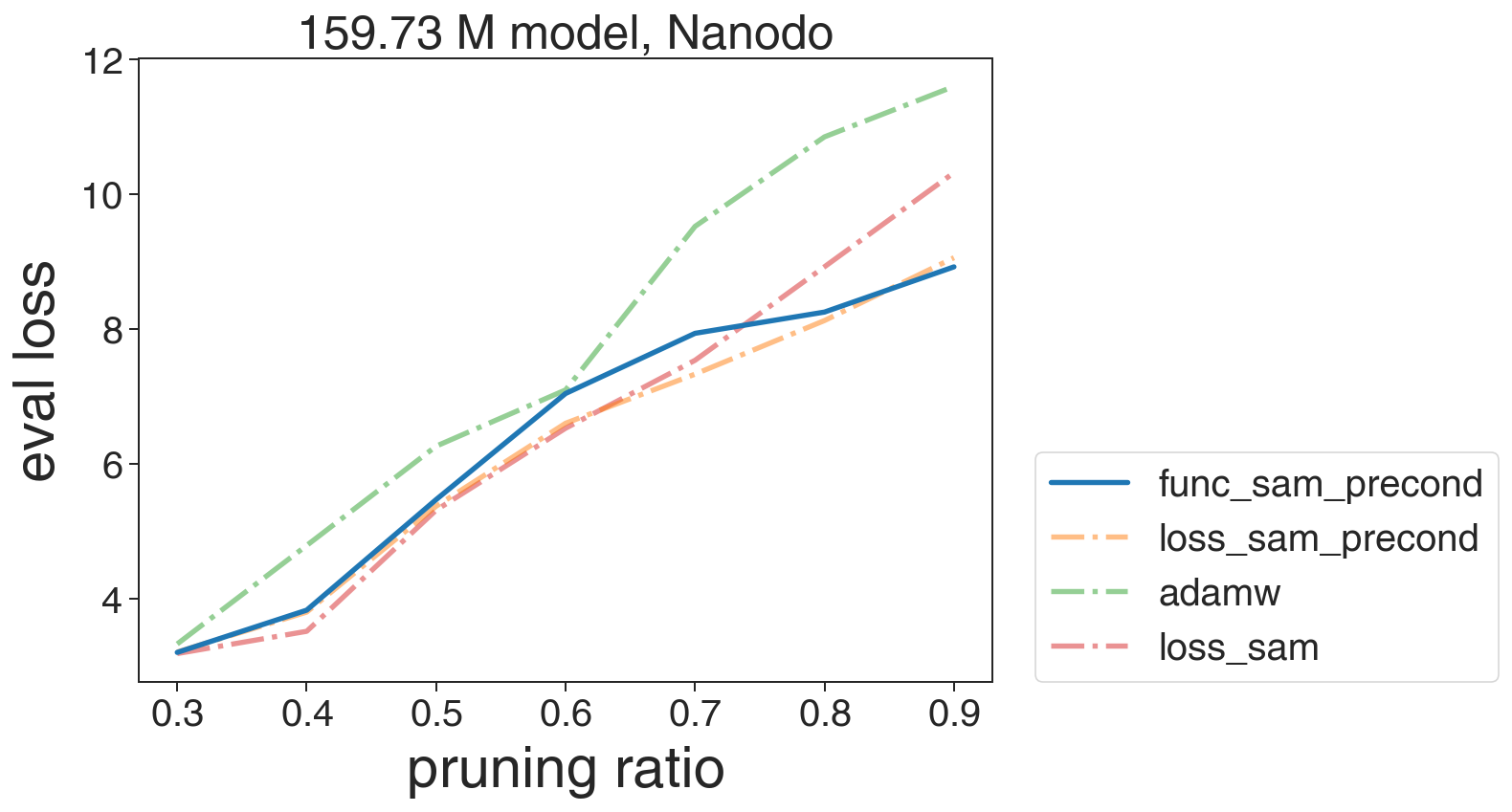}
\end{tabular}
\caption{\textit{Effect of one-shot (unstructured) global magnitude pruning:} We see that sharpness minimization methods tend to degrade more gracefully as increasing number of parameters are pruned. Also, from this figure we can see that the performance gained imparted by \funcSAM over \adamw is equivalent to setting about $25\%$ parameters of zero, and is thus significant.}
\label{fig:pruning}
\end{figure*}

\vspace{-1em}
\subsection{Architectural Details.}\label{app:archi}
The $23.9$M, $42.5$M, $117.9$M, and $1208$M models have the same depth of $6$, and whose width has been scaled together with the number of heads. In particular, these correspond to $h=9$, $12$, $20$, and $64$ heads per block and the width $m$ scales as $m=64\times h$, and the MLP dimension is $f=4\times m$. The $2$M model used for prototyping has depth $3$, $4$ heads per block, width $m=256$ and MLP dimension $f=1024$.\looseness=-1
\clearpage

\section{Additional theory}

\subsection{Argument for reducing matrix-vector products with inverse preconditioning}

\label{app:mvp-precond-argument}

We will use a random matrix model to reason about the effects of preconditioning with a matrix inverse. Random matrices have been used
to model the spectra of the Hessian of the large models found in machine learning, and in particular treating the Gauss-Newton and the functional Hessian/NME as
independent yields quantitative insights about the structure of the overall Hessian \citep{pennington17rmt}.

Consider two $N\times N$ symmetric
invertible matrices $\m{A}$ and $\m{B}$.
Suppose $\m{A}$ and $\m{B}$ are random and freely independent.
Free independence is the non-commutative analog to classical independence, implied by classical independence of entries in the limit of large matrix size \citep{pennington17rmt}.
The key feature it induces is
\begin{equation}
\mathbb{E}[\tr[\m{A}^{j}\m{B}^{k}]] = \mathbb{E}[\tr[\m{A}^{j}]]\mathbb{E}[\tr[\m{B}^{k}]]
\end{equation}
for any $j$, $k$, where $\tr$ is the trace.

Given a random unit vector $\m{v}$, the expected squared lengths of the matrix vector products with $\m{A}$ and $\m{B}$ are given by 
\begin{equation}
\mathbb{E}[\|\m{A}\m{v}\|^{2}] = \frac{1}{N}\mathbb{E}[\tr[\m{A}^{2}]],~\mathbb{E}[\|\m{B}\m{v}\|^{2}] = \frac{1}{N}\mathbb{E}[\tr[\m{B}^{2}]]
\end{equation}
The ratio $r_{1}$ of magnitudes is given by
\begin{equation}
r_{1} \equiv \frac{\mathbb{E}[\|\m{B}\m{v}\|^{2}]}{\mathbb{E}[\|\m{A}\m{v}\|^{2}]} = \frac{\mathbb{E}[\tr[\m{B}^{2}]]}{\mathbb{E}[\tr[\m{A}^{2}]]}
\end{equation}

Now consider the norm of $\m{w} = \m{A}^{-1}\m{v}$ passed through each matrix:
\begin{equation}
\mathbb{E}[\|\m{A}\m{w}\|^{2}] = \frac{1}{N},~\mathbb{E}[\|\m{B}\m{w}\|^{2}] = \frac{1}{N}\mathbb{E}[\tr[\m{A}^{-1}\m{B}^{2}\m{A}^{-1}]]
\end{equation}
The new ratio of magnitudes $r_{2}$ is given by
\begin{equation}
r_{2} \equiv \frac{\mathbb{E}[\|\m{B}\m{w}\|^{2}]}{\mathbb{E}[\|\m{A}\m{w}\|^{2}]} = \mathbb{E}[\tr[\m{A}^{-1}\m{B}^{2}\m{A}^{-1}]] = \frac{\mathbb{E}[\tr[\m{B}^{2}]]}{\mathbb{E}[\tr[\m{A}^{-2}]]^{-1}}
\end{equation}
From Jensen's inequality, $\tr[\m{A}^{2}]\geq \tr[\m{A}^{-2}]^{-1}$. Therefore $r_{2}>r_{1}$; preconditioning by the inverse of $\m{A}$
causes matrix-vector products to upweigh products with $\m{B}$ relative to products with $\m{A}$, relative to the unpreconditioned product.

In neural network settings, there are non-trivial relationships between the gradient, the Gauss-Newton matrix $\HG$ and the functional Hessian $\HF$; however,
the eigenvectors and eigenvalues of $\HG$ and $\HF$ are only weakly related. Therefore even though the exact calculations in the example
above don't hold, we suspect that generically preconditioning by $\HG^{-1}$ will downweigh $\HG\eps^{*}$ compared to $\HF\eps^{*}$.

\clearpage

\section{Code Snippets for \SAM and \funcSAM}\label{app:code}
\input{code}

\clearpage
\section{\angleSAM}\label{app:angle-sam}

In this section, we present a general variant of \SAM, which includes both \funcSAM and \SAM as its particular instantiations.  The core idea is that once we have been able to decompose the sharpness gradient into those arising from logit and functional paths, we can design our bespoke or custom versions of \SAM which lean more or less towards one path than another. 

To do so, let's assume that out custom path is at an angle $\phi$ with the functional path. We weigh the functional path by a factor of $\cos(\phi)$, while we weigh the logit path by $\sin(\phi)$. Then the gradient update in \angleSAM can be described as:

\begin{align}
     &\m{g}_{\, \angleSAM} := \nabla_\btheta \Loss(\btheta) + \sin(\phi) \,\cdot \, \m{g}_\logitText + \cos(\phi) \, \cdot \,\m{g}_\funcText \\
     & = \nabla_\btheta\Loss(\btheta) + \sin(\phi) \,\cdot \,\left[ \nabla_\btheta \boldsymbol{F}(\btheta)\cdot  \nabla_{\boldsymbol{F}}\Loss(\btheta+ \rad \, \beps^\ast)- \nabla_\btheta\Loss(\btheta)\right] + \cos(\phi) \,\cdot \, \left[\nabla_\btheta \boldsymbol{F}(\btheta+\rad\,\beps^\ast) \cdot  \nabla_{\boldsymbol{F}}\Loss(\btheta) - \nabla_\btheta\Loss(\btheta)\right]\\
     & = \nabla_\btheta\Loss(\btheta) + \rho \sin(\phi) \,\cdot \, \lcolor{\HG}\,\cdot\, \beps^\ast + \rho \cos(\phi) \,\cdot \, \fcolor{\HF}\,\cdot\, \beps^\ast  + \mathcal{O}(\rho^2)
\end{align}

We see that $\phi=\frac{\pi}{4}$ recovers \SAM upto first order in $\rho$, while $\phi=0$ would yield \funcSAM and $\phi=\frac{\pi}{2}$ would result in a variant that optimizes solely along the logit path and which can thus be called \logitSAM. \textit{All in all, this shows how \angleSAM is a clean generalization of \SAM, that equips it with a perturbation angle in addition to the usual perturbation radius $\rho$.}

At some level, this above formulation could be viewed as making an assumption that these two paths are orthogonal\footnote{This is not far-fetched.
As we noted in our experiments, these two paths tend to be anti-correlated, but often the correlation is quite small in magnitude and and approaches zero.}. On another level, one can simply think of these weights as a mere strategy to obtain convenient weight settings that have their sum of squares as unity.   

\textit{We expect that this approach might pay dividends in different model-dataset-optimizer triples, and we expect this to be an interesting direction for future work.}

\end{document}

%% file: dimensions.tex
\newcommand{\dataDim}{n}

%% file: math_defs.tex
\usepackage{nicematrix}

\newcommand{\mb}{\mathbf}
\newcommand{\m}[1]{{\boldsymbol{\mathbf{#1}}}}

\newcommand{\btheta}{{\pmb \theta}}
\newcommand{\bTheta}{{\pmb \Theta}}

\newcommand{\beps}{{\pmb \epsilon}}

\newcommand{\x}{\mathbf x}
\newcommand{\y}{\mathbf y}
\newcommand{\z}{\mathbf z}
\newcommand{\funcsym}{\mathrm{F}}

\newcommand{\ggnsym}{\mathrm{G}}

\newcommand{\Loss}{{\mathcal{L}}}

\makeatletter

\DeclareMathOperator*{\argmin}{argmin}

\DeclareMathOperator{\diag}{diag}
\DeclareMathOperator{\softmax}{softmax}

\newcommand{\tr}{\text{tr}}

\newcommand{\HG}{\mb{H}_{\ggnsym}}

\newcommand{\HF}{{\mb{H}_{\funcsym}}}

\newcommand{\HL}{{\mb{H}_{\Loss}}}

\newcommand{\D}{{\mathcal{D}}}

\newcommand{\ellbold}{\boldsymbol\ell}

\newcommand*{\colorboxed}{}
\def\colorboxed#1#{%
	\colorboxedAux{#1}%
}
\newcommand*{\colorboxedAux}[3]{%
	\begingroup
	\colorlet{cb@saved}{.}%
	\color#1{#2}%
	\boxed{%
		\color{cb@saved}%
		#3%
	}%
	\endgroup
}

\newcommand{\Reals}[1]{{\mathbb{R}^{#1}}} %

\newcommand{\pb}{{\mb p}}

\usepackage[scr=false]{rsfso}

\def\Mm{{\bf M}}

\usepackage{xcolor}%
\usepackage{ifmtarg}%
\usepackage{xifthen}%
\usepackage{environ}%
\usepackage{multido}%
\usepackage{lipsum}%
\usepackage{mdframed}
\usepackage{xspace}

\definecolor{colorFunc}{RGB}{200,104,16}

\definecolor{colorLogit}{RGB}{31,119,180}

\definecolor{coral}{HTML}{FF7F50}
\definecolor{peach}{HTML}{CC5500}
\definecolor{NavyBlue}{RGB}{0, 0, 128}
\definecolor{CrimsonRed}{RGB}{220, 20, 60}
\definecolor{ForestGreen}{RGB}{34, 139, 34}
\definecolor{Gold}{RGB}{204, 172, 0}
\definecolor{PaleRed}{rgb}{0.95, 0.3, 0.3}
\definecolor{MPLOrange}{rgb}{1.0, 0.6470588235294118, 0.0}
\definecolor{MPLGreen}{rgb}{0.0, 0.5019607843137255, 0.0}
\definecolor{ForestGreen}{RGB}{34,139,34}
\definecolor{OliveGreen}{RGB}{107,142,35}
\definecolor{PurplePrint}{RGB}{117, 112, 179}
\definecolor{GreenPrint}{RGB}{27, 158, 119}
\definecolor{RedPrint}{RGB}{217, 95, 2}
\definecolor{VeryLightGray}{rgb}{0.9,0.9,0.9}
\definecolor{samColor}{RGB}{214,38,40}
\definecolor{adamwColor}{RGB}{61,200,215}

\newcommand{\fcolor}[1]{\textcolor{colorFunc}{#1}}
\newcommand{\lcolor}[1]{\textcolor{colorLogit}{#1}}
\newcommand{\spcolor}[1]{\textcolor{black}{#1}}

\newcommand{\SAM}{{\textcolor{black}{\textsc{SAM}}}\xspace}

\newcommand{\adamw}{{\textsc{AdamW}}\xspace}

\newcommand{\funcSAM}{{\textcolor{black}{\textsc{Functional-SAM}}}\xspace}

\newcommand{\funcSAMbrief}{{\textsc{Func-SAM}}\xspace}
\newcommand{\precondSAM}{{\textsc{preconditioned SAM}}\xspace}
\newcommand{\precondSAMbrief}{{\textsc{precond SAM}}\xspace}

\newcommand{\penaltySAM}{{\textsc{penalty-SAM}}\xspace}
\newcommand{\precond}{{\textsc{precond}}\xspace}
\newcommand{\logitSAM}{{\textsc{Logit-SAM}}\xspace}
\newcommand{\angleSAM}{{\textsc{Angle-SAM}}\xspace}

\usepackage{thmtools}
\usepackage{cleveref}

\newtheorem{assump}{Assumption}

\makeatletter
\newcommand{\neutralize}[1]{\expandafter\let\csname c@#1\endcsname\count@}
\makeatother

%% file: template_content.tex
\section{Introduction}\label{intro}

One of the most fundamental questions in machine learning research is: how do we train models that are useful beyond  their training data?
This question arises in multiple scenarios --- from generalizing to unseen samples, dealing with distribution shift, and fine-tuning on specific domains.
A commonly held belief is that it is important for models to converge to \emph{well-behaved}
and \emph{robust} solutions. The `regularity' of the model is often obtained using \emph{regularization} techniques, which --- even in the day and age of LLMs --- remain an indispensable part of any training algorithm.

Some of the most prominent regularization methods include weight decay~\citep{krogh1991simple}, dropout~\citep{srivastava2014dropout}, data augmentation~\citep{ciregan2012multi,krizhevsky2012imagenet}, Mixup~\citep{zhang2017mixup}, and curvature-based controls~\citep{foret2020sharpness,wu2020adversarialweightperturbationhelps}.
In recent years, \emph{curvature regularization} techniques have gained popularity due to their effectiveness in promoting generalization. These techniques bias learning dynamics to areas of lower curvature (i.e., less sharp regions)
in the loss landscape~\citep{chaudhari2017entropysgdbiasinggradientdescent,keskar2017largebatchtrainingdeeplearning,foret2020sharpness,DBLP:journals/corr/abs-2006-07897,wu2020adversarialweightperturbationhelps}.
The origins of these curvature or \emph{sharpness minimization} techniques can be traced back to the classical ideas of minimum description length \citep{rissanen1978modeling,hinton1993keeping,Hochreiter1997FlatM}.
Lately, their development has been inspired by
their success in a large-scale correlational study~\citep{jiang2019fantasticgeneralizationmeasures} and in the NeurIPS generalization competition~\citep{jiang2020neurips2020competitionpredicting}.\looseness=-1

Sharpness minimization has demonstrated significant improvement on vision tasks. In contrast, there has not been much uptake
of these methods in NLP\footnote{The most notable exception of \SAM in NLP is in the fine-tuning scenario~\citep{bahri2022sharpnessawareminimizationimproveslanguage}, where the parameters are constrained to move smaller distances by the very nature of the setup.\looseness=-1} and, especially for as cornerstone~\citep{brown2020languagemodelsfewshotlearners} a task in NLP as language modeling. \mbox{Curiously} enough, we observe that Sharpness Aware Minimization (\SAM),
one of the best studied sharpness regularization methods \citep{foret2020sharpness}, shows poor performance here; indeed,
its validation metrics are typically worse than \adamw throughout training (Figure~\ref{fig:sam-adamw-loss-curve}), despite using
more computation per step.  Hence, this raises the following questions which form the basis of our work:

\begin{figure}[ht]
\begin{center}
\centerline{\includegraphics[width=0.95\columnwidth]{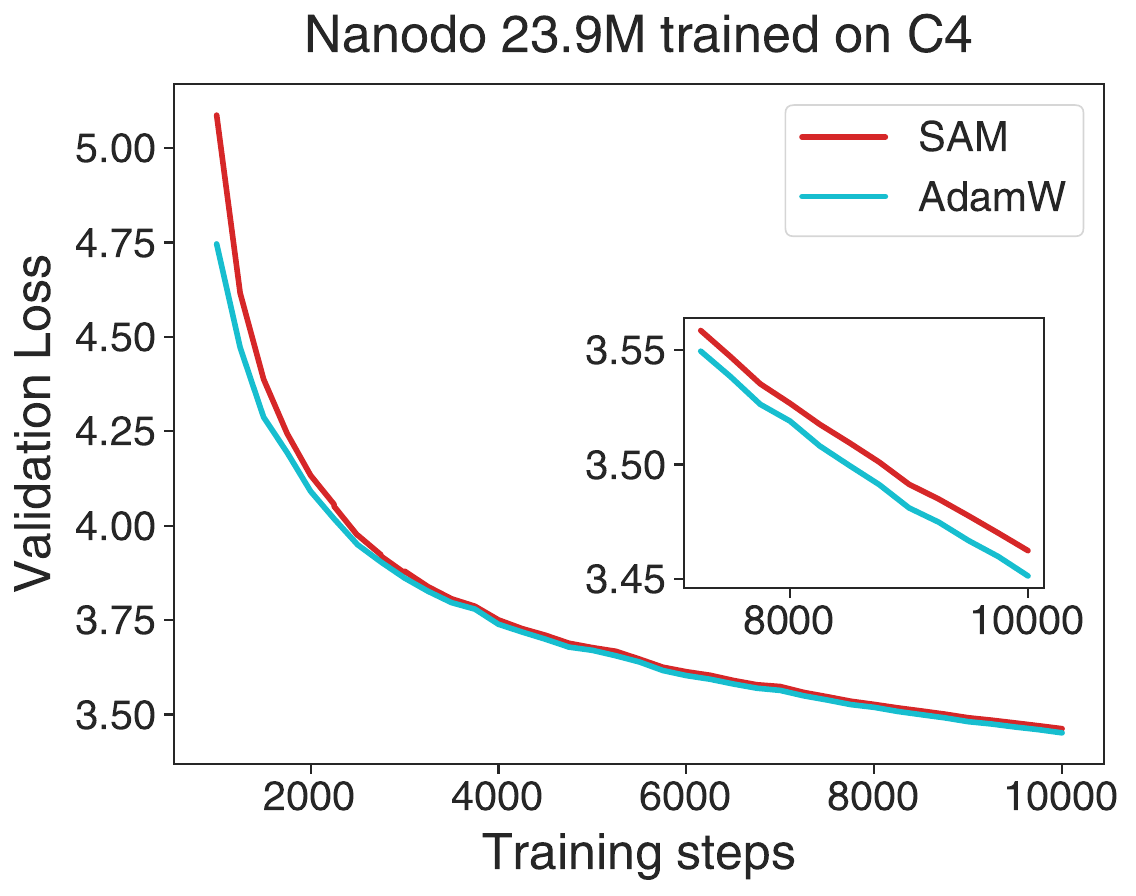}}
\caption{Evaluation loss curves of \textcolor{adamwColor}{\textsc{AdamW}} and \textcolor{samColor}{\textsc{SAM}} for Nanodo decoder-only Transformer model~\citep{nanodo} on the C4 dataset~\citep{raffel2020exploring}.\looseness=-1
}
\label{fig:sam-adamw-loss-curve}
\end{center}
\vskip -0.4in
\end{figure}
\begin{quote}
    \textit{What are the reasons for  \SAM's poor performance in language modeling? How can they be addressed to successfully yield generalization benefits, while being equal in cost as \SAM? }
\end{quote}
Towards this end,   in \textbf{Section~\ref{sec:sharpness-paths}}, we perform \textit{a novel analysis of the path \SAM takes to sharpness reduction}, and show that it can be split into two contributions --- one which
modifies the \emph{logit} statistics to reduce sharpness, and the other which modifies the geometry of the \emph{function} itself. We measure
these two contributions and find that in vision (where \SAM works well) the contributions are relatively balanced; in contrast,
in language modeling settings the logit path to sharpness minimization dominates.

 We hypothesize that the functional path to sharpness minimization needs to be amplified in the language setting, and in \textbf{Section~\ref{sec:algos}}, develop 
a \textit{novel sharpness minimization algorithm} called \funcSAM that achieves this. 
Additionally, we motivate another algorithm, \precondSAM, based on preconditioning, and give a theoretical argument that shows it  promotes the functional path.

In \textbf{Section~\ref{sec:empirics}}, we show that \funcSAM and \precondSAM provide improvements over baseline  \adamw when trained on the C4 dataset using the Nanodo Transformer
codebase~\citep{nanodo}. Moreover, \funcSAM and \precondSAM can be combined to yield maximal gains.
This resulting combination consistently improves validation metrics in both fixed number of steps as well as Chinchilla-like scaling settings~\citep{hoffmann2022trainingcomputeoptimallargelanguage} \textit{at a variety of model sizes, spanning three orders of magnitude.} 

We conclude with an extensive discussion, in \textbf{Section~\ref{sec:disc},} of additional \SAM variants that may be useful in other domains as well as important future directions implied by our work.\looseness=-1

\section{Setup and Background}

Let us assume that we are given data points $\z\in\mathcal{Z}$ drawn i.i.d. from some (unknown) distribution $\D$, where the samples $\z$ are input-output tuples $(\x, \y)$, with the input $\x \in \mathbb{R}^d$ of dimension $d$ and the targets $\y \in \mathbb{R}^K$ of dimension $K$. We seek to model the input-output relation via a neural network $\boldsymbol{f}_\btheta(\x) :\mathbb{R}^d \mapsto \mathbb{R}^K$ with learnable parameters $\btheta\in\mathbb{R}^p$, such that $\boldsymbol{f}_\btheta(\x) \approx \y \;, \, \forall \, (\x, \y) \in \D$. We take the usual route of empirical risk minimization~\citep{vapnik1991principles} and consider $\btheta^\ast:=\argmin_{\btheta\in\bTheta} \Loss(\btheta)$ with $\Loss(\btheta) := \frac{1}{\dataDim} \sum_{i=1}^{\dataDim} \ellbold(\z_i; \btheta)$ and where, $S=\lbrace\z_i\rbrace_{i=1}^n$ is the training set of size $n$ and $\ellbold$ denotes the loss function. Hereafter we will consider the loss to be cross-entropy, which is the most popular choice; however, our analyses extend to other loss functions as well.

Under the above setup,~\citet{foret2020sharpness} formulates sharpness-aware minimization as the following min-max problem:
$\min_\btheta \max_{\|\beps\| \leq \rho} \, \Loss(\btheta + \beps)\,,$ where $\beps$ denotes a perturbation of the parameters. In particular, the inner maximization is approximated to first order in the perturbation, 
\begin{align}\label{eq:sam}
\vspace{-2mm}
\max_{\|\beps\| \leq \rho} \, \Loss(\btheta) \,+\, \beps^\top \nabla_\btheta \Loss(\btheta)\,,
\end{align}
which subsequently yields $\beps^\ast(\btheta) = \rho\,{ \nabla_\btheta \Loss(\btheta)}/{\|\nabla_\btheta \Loss(\btheta)\|} $  as the optimal perturbation. In \SAM~\citep{foret2020sharpness}, the authors propose making an update along the direction,
\begin{equation}\label{eq:sam-update}
\m{g}_{\, \SAM} = -\nabla_\btheta\Loss(\th+\rad\, \eps^*),~\eps^*\equiv \frac{\nabla_{\th}\Loss(\th)}{\|\nabla_{\th}\Loss(\th) \|}\,.
\end{equation}

\section{The Dual Routes to Sharpness Minimization}\label{sec:sharpness-paths}

In this section, we develop a diagnostic tool to understand the failures of \SAM in language modeling settings. Our theoretical analysis shows
that there are two possible routes to sharpness minimization via \SAM --- the \emph{logit} path and the \emph{functional} path.
We derive quantities which can be used to measure the extent to which each route is active.
We find that each path is relatively balanced in vision, but in language modeling settings (where \SAM performs poorly) the logit path overwhelms 
the functional path.

\subsection{The Penalty Formalization}

We begin by looking at \penaltySAM, an alternative version of \SAM that is often used in 
theoretical analyses~\citep{andriushchenko2022understandingsharpnessawareminimization,dauphin2024neglected} and whose gradient matches $\m{g}_{\, \SAM}$ to first order in
$\rad$:
\begin{align}\label{eq:pen-sam}
    \min_\btheta\, \Loss(\btheta) + \textcolor{black}{\underbrace{\rho \;\|\nabla_\btheta \Loss(\btheta)\|}_{\spcolor{\SP}}}\,,
\end{align}
We can think of the added gradient norm term as a sharpness penalty $\spcolor{\SP}$. To understand how $\spcolor{\SP}$ influences optimization, we investigate the structure of its gradient:
\begin{align}\label{eq:gradnorm-grad}
\spcolor{\nabla_\btheta \, \SP(\btheta)}:=\rho \, \frac{\partial }{\partial\btheta} \|\nabla_\btheta \Loss(\btheta)\|&=   \left(\frac{\partial }{\partial\btheta}\nabla_\btheta \Loss(\btheta)\right) \cdot \beps^\ast(\btheta)
\,.
\end{align}
The gradient itself can be decomposed by the chain rule as
\begin{align}
    \nabla_\btheta \Loss(\btheta) :=  \nabla_\btheta \boldsymbol{F}(\btheta) \cdot \nabla_{\boldsymbol{F}} \Loss(\btheta)\,,
\end{align} where $\boldsymbol{F}(\btheta) = \begin{pmatrix} \boldsymbol{f}_\btheta(\x_1)^\top \cdots \boldsymbol{f}_\btheta(\x_n)^\top\end{pmatrix}^\top \in \Reals{Kn}$ collates the output over the entire dataset.
The first term $\nabla_\btheta \boldsymbol{F}(\btheta)$ represents the Jacobian (derivative of the model function outputs with
respect to parameters); while the second term $\nabla_{\boldsymbol{F}} \Loss(\btheta)$ is the derivative of the loss with respect to the outputs and which comes out to be the difference of the (softmax-ed) logits and the targets.

Using the product rule, we
can rewrite Eqn.~\ref{eq:gradnorm-grad} as, 
\begin{align}\label{eq:sharp-grad-comp}
&     \underbrace{\left\lbrack \nabla_\btheta \boldsymbol{F}(\btheta)\cdot \lcolor{\frac{\partial  }{\partial \btheta} \big(\nabla_{\boldsymbol{F}} \Loss(\btheta)\big)} \right\rbrack\,\,\cdot \beps^\ast(\btheta)}_{\spcolor{\SP} \text{ gradient from \textbf{\lcolor{logit}} perturbation}\,:=\,\lcolor{\glog}} \, + \,\nonumber\\
    &   \underbrace{\left\lbrack\fcolor{\frac{\partial  }{\partial \btheta} (\nabla_\btheta \boldsymbol{F}(\btheta))}\cdot \nabla_{\boldsymbol{F}} \Loss(\btheta)\right\rbrack\cdot \beps^\ast(\btheta)}_{\spcolor{\SP} \text{ gradient from perturbing the \fcolor{\textbf{function}} \textbf{\fcolor{Jacobian}}}\,:=\,\fcolor{\gfun}}\,.
\end{align}
The first term $\lcolor{\glog}$ represents the \emph{logit-path} to sharpness minimization --- directly optimizing the sharpness via the effect
of the logits on the loss.
In contrast, the second term $\fcolor{\gfun}$ represents the \textit{functional-path} to sharpness minimization --- that is, sharpness
minimization via modification of the Jacobian statistics.

\subsection{Contribution of \lcolor{Logit} and \fcolor{Functional} paths}
We can explicitly relate the logit and functional paths to sharpness minimization ($\lcolor{\glog}$ and $\fcolor{\gfun}$ in Eqn. ~\ref{eq:sharp-grad-comp})
with a decomposition of the Hessian of the loss $\spcolor{\HL}=\nabla_\btheta^2 \Loss$,
\begin{align}\label{eq:pen-grad}
    \spcolor{\nabla_\btheta \, \SP}&= \lcolor{\glog} + \fcolor{\gfun} = {\lcolor{\HG} \cdot \beps^\ast} \, +\, {\fcolor{\HF}\cdot  \beps^\ast} = \spcolor{\HL} \cdot \beps^\ast
\end{align}
where, we use the Gauss-Newton Decomposition of the Hessian~\citep{Schraudolph2002FastCM}, namely:
\begin{align}\label{eq:hess}
    \spcolor{\HL} =  \lcolor{\HG} &+ \fcolor{\HF} =  \frac{1}{\dataDim}\sum_{i=1}^{\dataDim} \nabla_\btheta \boldsymbol{f}_\btheta(\x_i) \lcolor{\big[\nabla^2_{\boldsymbol{f}} \,\ellbold_i\big]} \, \nabla_\btheta \boldsymbol{f}_\btheta(\x_i)^\top\nonumber\\
    &+ \frac{1}{\dataDim}\sum_{i=1}^{\dataDim} \,
    {\sum_{k=1}^K \,\, [\nabla_{\boldsymbol{f}}\ellbold_i]_k\,}  \cdot \,\fcolor{\nabla^2_\btheta \,\boldsymbol{f}^{\,{k}}_\btheta(\x_i)}\end{align}
where, the \lcolor{Generalized Gauss Newton (GGN)} $(\lcolor{\HG})$ and the \fcolor{functional Hessian} $(\fcolor{\HF})$~\citep{singh2021analytic,10.5555/3618408.3619732} are the component matrices of the loss Hessian $(\spcolor{\HL})$.
The GGN term captures the curvature of the linearized model; in contrast, the functional Hessian (also known as the Nonlinear Modeling
Error \citep{dauphin2024neglected}) captures curvature due to model second derivatives.

Thus, we can interpret the
\textit{gradient through the logit and functional paths based on the  Hessian component they depend upon} --- the
GGN and functional Hessian respectively.\looseness=-1

\textbf{Normalized composition of Sharpness Gradient.} To better gauge which mode of sharpness gradient dominates, we will measure the following natural quantities, $\lcolor{\tlog}, \fcolor{\tfn}, \tc$, where the three sum to $1$:
\begin{equation}\label{eq:norm-grad-contrib}
\begin{split}
    &\lcolor{\tlog} =  \frac{\|\lcolor{\glog}\|^2}{\|\spcolor{\nabla_\btheta \, \SP}\|^2}, \fcolor{\tfn} = \frac{\|\fcolor{\gfun}\|^2}{\|\spcolor{\nabla_\btheta \, \SP}\|^2},\\
    &\tc = 2\, \frac{\langle\lcolor{\glog}, \fcolor{\gfun}\rangle}{\|\spcolor{\nabla_\btheta \, \SP}\|^2}
    \end{split}
\end{equation}
Hence based on their values, \textit{we can realize whether sharpness minimization will prioritise reduction of logit sharpness more or that of functional sharpness}, as well as how correlated they are by looking at the cross term.

\begin{figure*}[h]
\centering
\begin{tabular}{cc}
\includegraphics[height=0.3\linewidth]{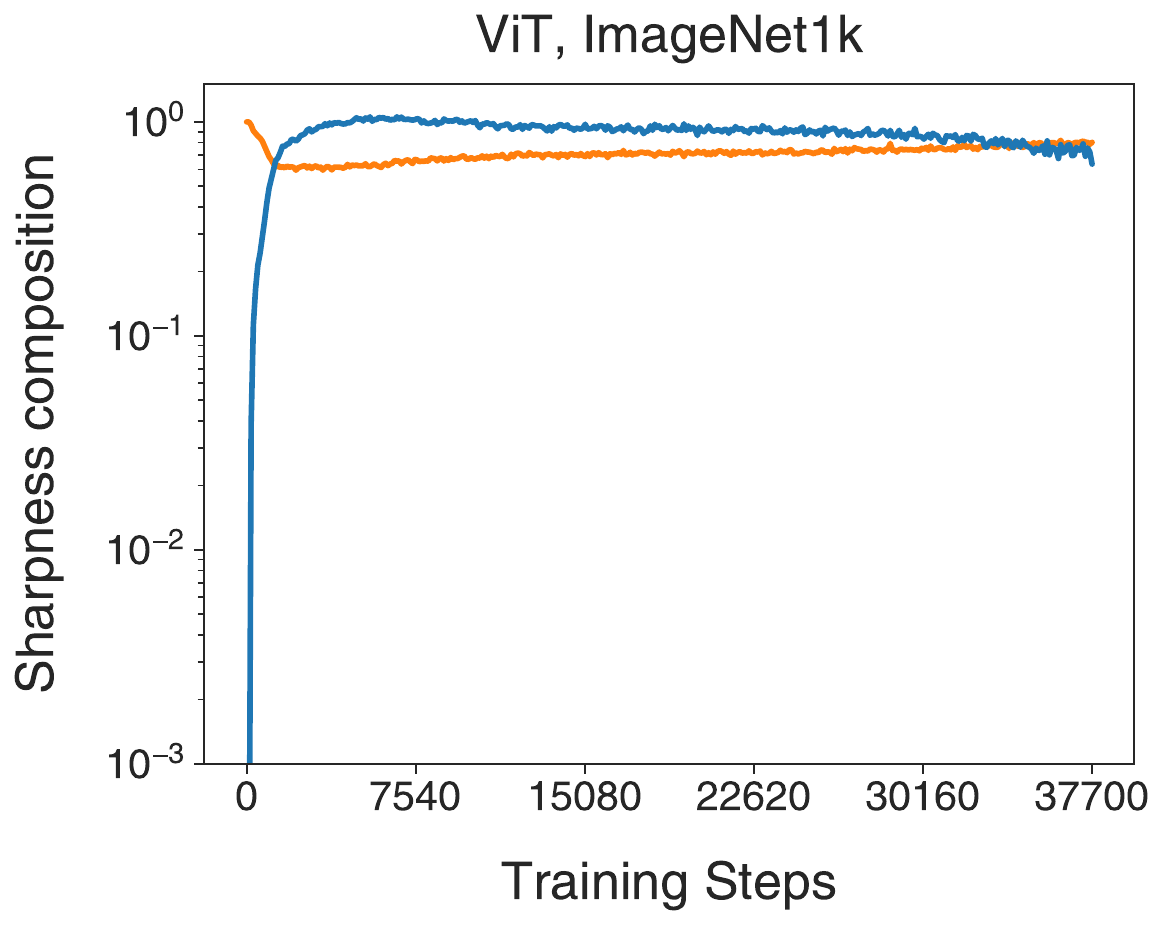} \hfill& \includegraphics[height=0.3\linewidth]{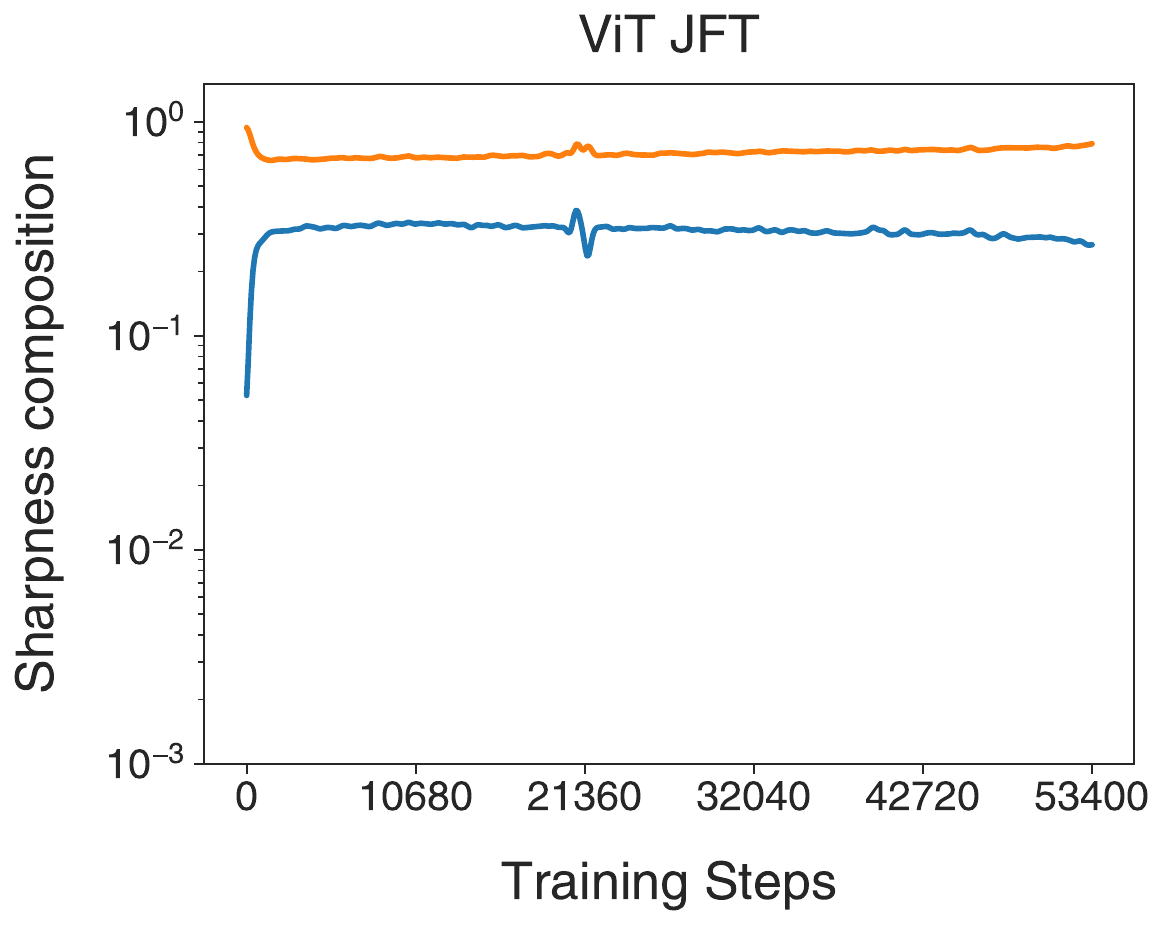} \\
\includegraphics[height=0.3\linewidth]{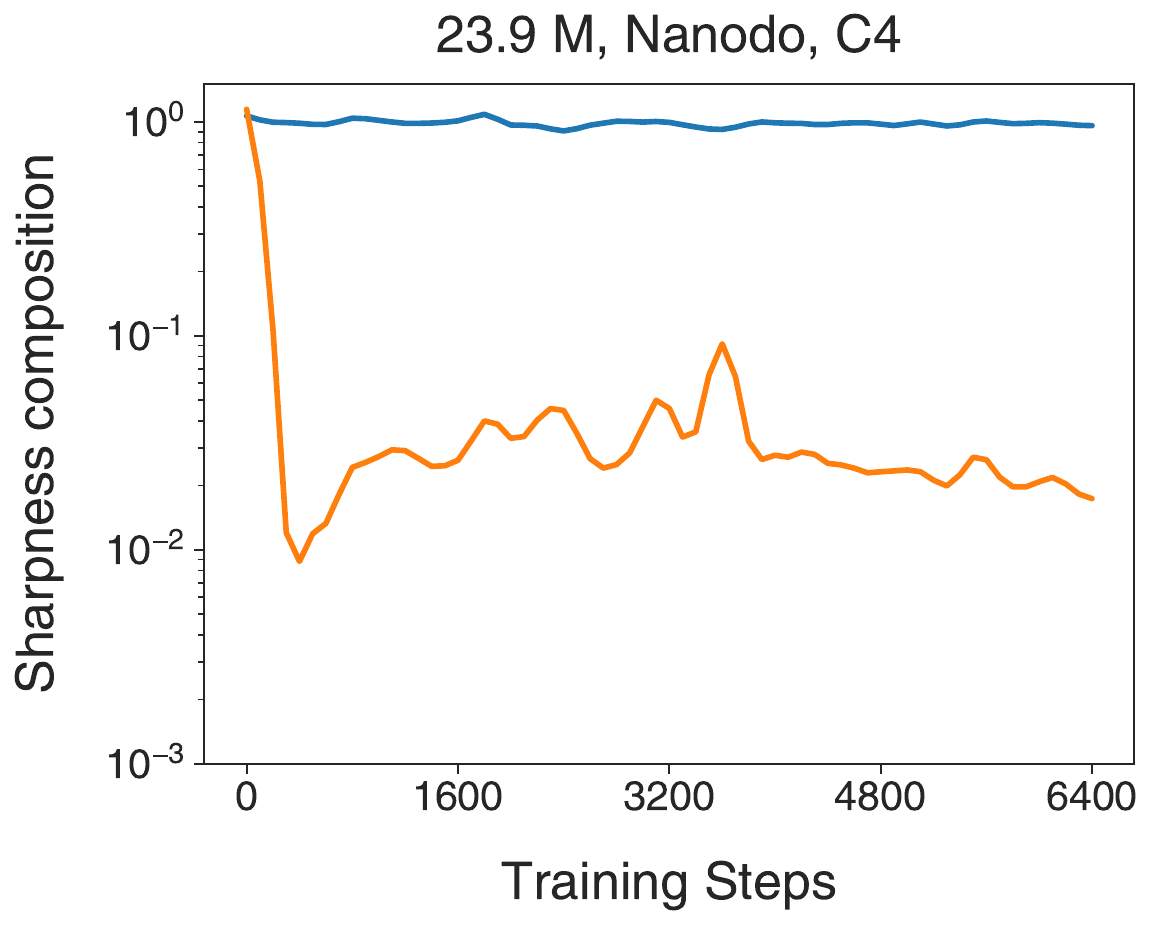} \hfill& \includegraphics[height=0.3\linewidth]{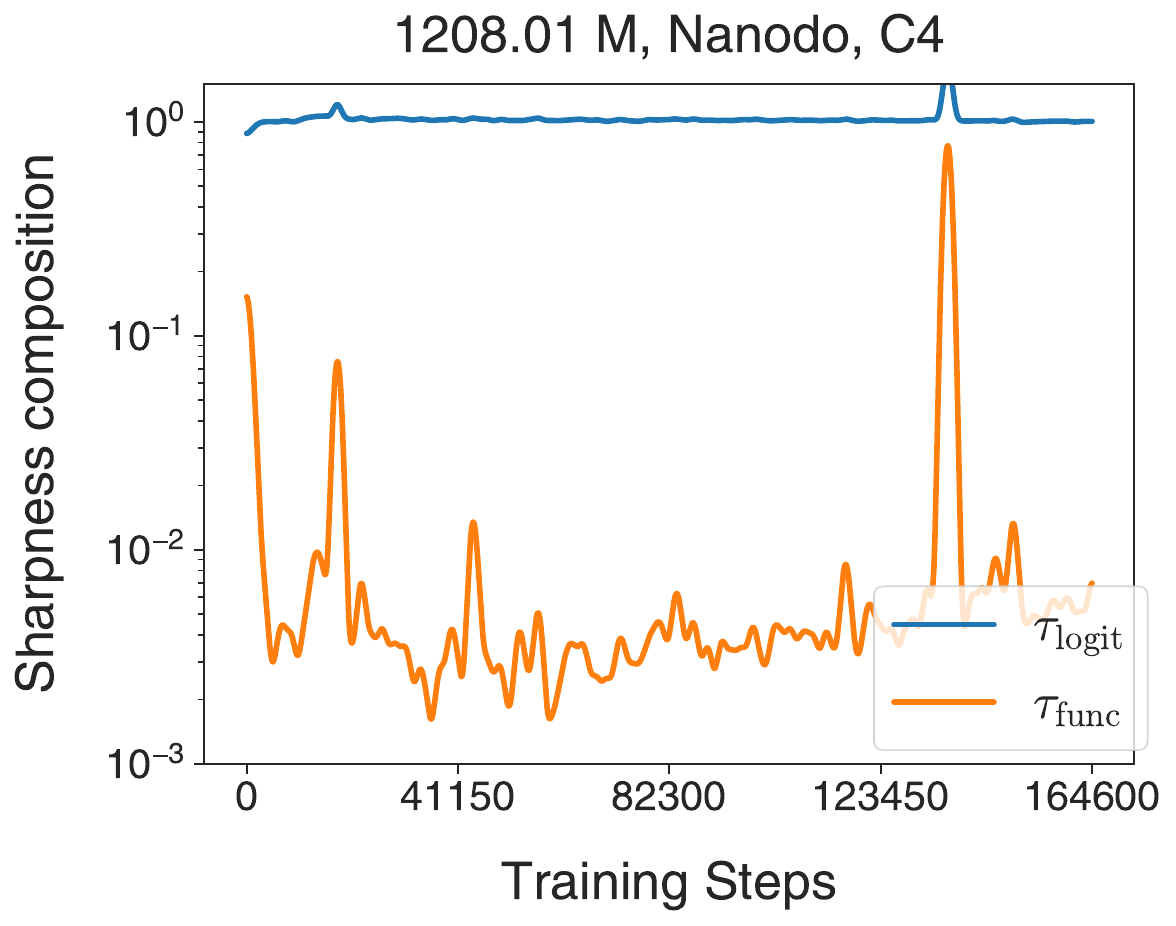}
\end{tabular}

\caption{Normalized sharpness contributions $\lcolor{\tlog}$ and $\fcolor{\tfn}$ show dramatically different trends across modalities. For ViT trained on ImageNet-1K (top left) and JFT (top right), $\lcolor{\tlog}$ starts near $0$ but quickly increases to a comparable magnitude as $\fcolor{\tfn}$. For Transformer models trained on C4 (bottom left and bottom right), $\lcolor{\tlog}\gg \fcolor{\tfn}$ after the first few steps of training. This suggests that the pathways to sharpness regularization are more imbalanced in NLP compared to vision settings, which may contribute to the poor performance of \SAM in NLP settings. $\tc$ (plotted in Appendix \ref{app:sharp_plots}) is usually negative, suggesting the two methods of sharpness regularization tend to be antagonistic.}
\label{fig:sam_contribution_figure}
\end{figure*}

\subsection{Composition of Sharpness Gradient in Practice}
We will now analyze \SAM's behavior in language modeling, focusing on how the sharpness gradient composition reveals key differences between its application in language and vision tasks. More concretely, we consider next-token prediction task in the case of language using the C4 dataset and image classification in the case of vision. Since the typical vocabulary sizes in language is in the order of tens of thousands, so besides ImageNet-1K, we adopt other datasets like  JFT~\citep{sun2017revisitingunreasonableeffectivenessdata} and ImageNet-21K~\citep{ridnik2021imagenet21kpretrainingmasses} to make the settings further comparable in terms of number of outputs. For both settings, we employ Transformer-based networks, Nanodo~\citep{nanodo}, which is a simplified version of GPT-2~\citep{radford2019language}, in language modeling and Vision Transformer (ViT,~\citealp{dosovitskiy2021imageworth16x16words}) for vision tasks. Furthermore, in both cases, we train with \adamw as the optimizer and measure the normalized sharpness gradient contributions (Eqn.~\ref{eq:norm-grad-contrib}) throughout training using exact Hessian-vector products. We present the results for Nanodo, C4 as well as ViT with ImageNet-1K and JFT in Figure~\ref{fig:sam_contribution_figure}, and that on ImageNet-21K in Appendix \ref{app:sharp_plots}.

\textbf{Observations.} Comparing these figures, we find a stark contrast between the language and vision settings.
For vision, we find that the $\lcolor{\tlog}$ starts close to $0$
but that $\lcolor{\tlog}$ and $\fcolor{\tfn}$ quickly become comparable for most of the training process 
(Figure \ref{fig:sam_contribution_figure}, top left and top right).
In contrast for language, the logit-related gradient fraction $\lcolor{\tlog}$ is close to $1$ while the sharpness gradient related to the functional part is much smaller 
Figure~\ref{fig:sam_contribution_figure} (bottom left and bottom right). 
In all cases, $\tc$ tends to be negative through most of training (see Figure \ref{fig:sam_contribution_app} in the Appendix).
This suggests that the two paths to sharpness regularization are antagonistic --- taking one path moves you against the other.
In language modeling, the dominance of the logit path, combined with negative $\tc$, means that the overall
contribution from the gradient of $\spcolor{\SP}$ is unaligned, or even \emph{anti-aligned}, with the functional path to
sharpness.

\textbf{\lcolor{Logit Sharpness} vs \fcolor{Functional Sharpness}.} The above observations suggest that in NLP settings, \SAM is heavily biased towards minimizing sharpness
by reducing the gradient of the loss with respect to the logits. There is not much sharpness reduction happening due to making the function more well-behaved. While both of these contribute to decreasing sharpness, the underlying mechanisms show interesting differences. Recalling the relation of logit-sharpness to the Gauss-Newton term $\lcolor{\HG}$, we see that a simple but spurious way to decrease it is by make the network over-confident about its predictions. This is because of the presence of the term $\lcolor{\nabla^2_{\boldsymbol{f}} \,\ellbold_i}$, which equates to $\diag(\pb_i) - \pb_i\pb_i^\top$, where $\pb_i = \softmax(\lcolor{\boldsymbol{f}_\btheta(\x_i)})$. And thus as $\pb_i$ becomes more one-hot (irrespective of leading to the correct output or the incorrect), the logit sharpness will get reduced. 

In contrast, the functional sharpness is connected to the functional Hessian, and the ways of decreasing it (such as via modeling the target better $\pb_i \rightarrow \y_i$; or reducing the second-derivative of the function with respect to the parameters $\|\fcolor{\nabla^2_\btheta \boldsymbol{f}}\|\rightarrow 0$ and hence encouraging a functional simplicity) are  intuitively more desirable, even though setting a particular proportion of it might be unclear. 

These observations lead to the \textit{hypothesis that if we could encourage a reduction of functional sharpness in lieu of logit sharpness, then sharpness minimization might in fact work for language modeling too. }In the next section, we operate under this hypothesis and derive simple but principled modifications of the \SAM update rule.

\section{Algorithms to promote the functional path to sharpness minimization}\label{sec:algos}

From the \penaltySAM formulation in Eqn.~\ref{eq:pen-sam} and the form of the corresponding sharpness gradients in Eqn.~\ref{eq:pen-grad}, we can see that the functional path to sharpness can, in principle, be amplified by manipulating $\lcolor{\glog}$ and $\fcolor{\gfun}$.
However, while \penaltySAM is competitive with the \SAM algorithm, it is less robust than \SAM due to ill-behaved second derivatives
\cite{dauphin2024neglected}. In contrast to \penaltySAM, the perturbation approach in \SAM does not explicitly involve second-derivatives in the gradient. Therefore, it is important to develop strategies that promote the functional path using the methodology taken in \SAM.\looseness=-1 %

In this section, we discuss two strategies --- one direct and the other indirect --- 
that promote the functional path to sharpness
minimization, using algorithms which maintain the benefits of the \SAM formulation. 

\subsection{$\,$\funcSAM}

The simplest way to promote the functional path would be to use a perturbation which aligns more with $\fcolor{\gfun}$ than $\lcolor{\glog}$. We can find
such a perturbation by decomposing the \SAM update rule from Eqn.~\ref{eq:sam-update}, as follows:
\resizebox{0.5\textwidth}{!}{\begin{minipage}{0.5\textwidth}{
\begin{align}\label{eq:discrete-sam}
&\nabla_{\btheta}\Loss(\btheta+\rad\,\beps^\ast) = \underbrace{\left[ \nabla_\btheta \boldsymbol{F}(\btheta)\cdot  \lcolor{\nabla_{\boldsymbol{F}}\Loss(\btheta+ \rad \, \beps^\ast)}-\nabla_{\btheta}\Loss(\th)\right]}_{\lcolor{\glog} \;\text{up to first order in} \; \rho}
\nonumber\\[2mm]
&  +\underbrace{\left[\fcolor{\nabla_\btheta \boldsymbol{F}(\btheta+\rad\,\beps^\ast)} \cdot  \nabla_{\boldsymbol{F}}\Loss(\btheta)-\nabla_{\btheta}\Loss(\th)\right]}_{\fcolor{\gfun} \;\text{up to first order in} \; \rho}\,
+\, \nabla_{\btheta}\Loss(\th)\, +\,O(\rad^{2}) \end{align}
}
\end{minipage}
}

The first term is $\lcolor{\glog}$ up to first order in $\rad$, and the second term is $\fcolor{\gfun}$ up to first order in $\rad$. This suggests the following
update rule which we call \funcSAM:
\begin{align}\label{eq:func-sam-grad}
    \m{g}_{\, {\funcSAMbrief}} = -\fcolor{\nabla_\btheta \boldsymbol{F}(\btheta + \rho \,\beps^\ast)}\,\cdot\, \nabla_{\boldsymbol{F}}\Loss(\btheta)
\end{align}
where, we discard the $\lcolor{\glog}$ contribution.
Further, \funcSAM uses the same perturbation $\beps^\ast$ as the \SAM formulation, but \emph{only} perturbs the Jacobian --- thus
emphasizing the functional path to sharpness as desired. This update rule can be implemented as efficiently as regular \SAM using the 
very same vector-Jacobian product operations that are used to compute
gradient in most autodifferentiation frameworks. In Appendix~\ref{app:code}, we provide the JAX code snippets for \SAM and \funcSAM, demonstrating that the difference in their implementation is a matter of a few lines. Further, like \SAM, \funcSAM remains compatible with methods like \adamw which take a gradient and then further process it.

Overall, the update rule in Eqn.~\ref{eq:func-sam-grad} has the same cost as the \SAM update rule, and keeps the benefit of the finite-differences based perturbation approach to sharpness
estimation that keeps \SAM robust --- meeting all our initial goals.\looseness=-1

\subsection{$\,$\precondSAM}
In language modeling, Transformers~\citep{vaswani2017attention} are almost exclusively trained with \adamw or other adaptive methods, and SGD-based training is known to be significantly worse~\citep{liu2020understanding}. This is commonly attributed to the presence of heterogeneity in the gradients~\citep{liu2020understanding,noci2022signal,pan2023toward} and the curvature~\citep{zhang2024transformersneedadamhessian,ormaniec2024doesmeantransformerinsights,jiang2024does} across different ``modules'' (layer types, layers at different depths, etc.),
with the idea that \adamw alleviates heterogeneity and improves conditioning.

Naively combining \SAM and \adamw creates a potential mismatch ---
the SAM perturbation is carried out with respect to the unpreconditioned geometry. We hypothesize that this mismatch offers
grounds for spurious sharpness minimization.
To rectify this, we consider preconditioning the perturbation $\beps^\ast$ with the inverse of the same second-moment statistics $\textcolor{violet}{\Mm}$ from \adamw, resulting in the update:
\begin{align}\label{eq:precond-sam-update}
  & \m{g}_{\,{\precondSAMbrief }} \equiv - \nabla_\btheta \Loss\left(\btheta + \rho\, \textcolor{violet}{\Mm^{-1}(\btheta)} \, \beps^\ast(\btheta)\right)\,,
\end{align}
where $\m{g}_{\,{\precondSAMbrief }}$ is then passed to the rest of Adam in lieu of the gradient $\nabla_\btheta\Loss(\btheta)$. 

Another motivation for \precondSAM is that the gradients often align with the principal eigenspaces of $\HG$~\citep{gur2018gradient}. This would amplify
$\lcolor{\glog}$ over $\fcolor{\gfun}$, since $\lcolor{\glog} = {\HG}\,\cdot\,\beps^\ast$ and $\beps^*$ is parallel to the gradient.
Preconditioning $\beps^\ast$ by $\HG^{-1}$ reduces this effect,
thereby \emph{promoting the functional
path over the logit path} (see Appendix~\ref{app:mvp-precond-argument} for a more detailed argument).
Although $\HG^{-1}$ would be the best preconditioner in this regard, it is expensive to estimate generally and
\adamw already gives us a diagonal estimator in its own preconditioner $\textcolor{violet}{\Mm^{-1}}$ at no additional cost.

To conclude, this algorithm (Eqn.~\ref{eq:precond-sam-update}) has only marginally higher computational cost than standard \SAM + \adamw, 
and provides an indirect way to improve the contribution of the
functional path to sharpness minimization.

\section{Empirical Evaluation}\label{sec:empirics} 
\subsection{Setup}
\textbf{Training Lengths}. In order to evaluate language models of multiple sizes, we consider (pre-)training them in two scenarios: (a) when the training length is kept fixed across scales and (b) when the training length is adjusted as per compute-optimality considerations~\citep{kaplan2020scalinglawsneurallanguage,hoffmann2022trainingcomputeoptimallargelanguage}. 

\textit{(a) Fixed-Length Training Scheme.} 
In the former, we train all the models for 10K steps, which amounts to seeing roughly a total of $1.3$ billion tokens. The results for this setting are presented in Table~\ref{tab:fixed-10k}. \textit{(b) Chinchilla-style Training Scheme.}
Unlike traditional image-classification based scenarios,  training language models often involves determining compute-optimal scaling laws, and whereby models are trained on a corpus size in proportion to their parameters. Specifically, we follow the $20\times$ over-training policy suggested in the Chinchilla~\citep{hoffmann2022trainingcomputeoptimallargelanguage} training regime, and use it together with the fixed-depth and globally tuned learning rate scheme, as considered in~\citet{everett2024scaling}. The corresponding results are shown in Table~\ref{tab:chincilla}.

\textbf{Training Details and Hyperparameters}. In either scenario, we consider a batch size of $256$ sequences, of maximum length $512$, and evaluate model at $5$ different sizes: $2$M (for prototyping), $23.9$M, $42.5$M, $117.9$M, and $1208$M in terms of non-embedding parameters (see details in Appendix~\ref{app:archi}), and trained with \adamw~\citep{kingma2017adammethodstochasticoptimization,loshchilov2019decoupledweightdecayregularization} as the underlying optimizer on the C4 dataset~\citep{raffel2020exploring}. We tune the perturbation size $\rho$ separately for each method and model size. The learning rate and weight decay for \adamw have been tuned across model sizes yielding the values of $0.001, 0.1$ respectively. We use the Nanodo~\citep{nanodo} framework to implement these minimal decoder-only Transformer models, in Flax~\citep{flax2020github} together with JAX~\citep{jax2018github}. \looseness=-1

\subsection{Comparison of direct and indirect approaches}

Before carrying out an extensive evaluation across different model scales, we do initial prototyping on a smaller model size. This lets us save computational resources and scale up the most promising methods. In particular, we are interested in knowing which methods out of direct and indirect approaches, and their combinations, are most relevant. In Table~\ref{tab:precond}, we present results on a model with $2$M non-embedding parameters trained in the fixed-length regime. 

\begin{table}[ht!]
\centering
\caption{\textit{Effect of Preconditioning the SAM perturbation on \funcSAMbrief with $2M$ parameter model. } Lower is better.\looseness=-1}
\label{tab:precond}
\renewcommand{\arraystretch}{1.1} %
\setlength{\tabcolsep}{6pt} %

\begin{tabular}{@{}lc@{}}
\toprule
\textsc{Method}         & \textsc{Eval Loss} \\[1mm] \midrule
\adamw          & 3.90    \\ %

\SAM &  3.91 \\ \midrule
\precond\, \SAM       & 3.89    \\ 
\funcSAMbrief & 3.88    \\
\precond\,\funcSAMbrief & \textbf{3.86}    \\
\bottomrule
\end{tabular}

\end{table}

We notice that the plain version of \funcSAM outperforms both \adamw and preconditioned \SAM. 
But, \funcSAM can be further improved by using preconditioning alongside, yielding a significant improvement in terms of evaluation loss. Such second-decimal differences in evaluation loss are typical of the gains provided by new optimization methods for pre-training LLMs look like, e.g., in SOAP~\citep{vyas2024soap}. Moreover, it should be noted that the standard deviation across different seeds is somewhere in range of the fourth decimal place. Thus, the kind of gains shown in Table~\ref{tab:precond} are indeed quite significant.

While understanding the exact interplay of preconditioning with \funcSAM requires further study, the fact that their improvements accumulate suggests that \funcSAM also benefits from preconditioning through reducing heterogeneity across the modules and alleviating the mismatch between inner and outer optimization geometries in \SAM.
Hereafter, we will use the preconditioned variant for \funcSAM, instead of just the plain \funcSAM, to reduce the computational costs associated with testing at larger model scales.

\subsection{Results at Multiple Model Sizes}
\begin{table}[ht]
\caption{Evaluation loss comparison of different methods in a \textit{fixed-length (10K steps) training setup}. Lower is better.}
\renewcommand{\arraystretch}{1.2} %
\setlength{\tabcolsep}{1.5pt} %
\label{tab:fixed-10k}
\centering
\begin{tabular}{@{}lcccc@{}}
\toprule
\textsc{Size} &\begin{tabular}[c]{@{}c@{}}\precond\\   \funcSAMbrief\end{tabular} & \begin{tabular}[c]{@{}c@{}}\precond \\ \SAM\end{tabular} & \SAM  & \adamw  \\ \midrule
23.9 M  & \textbf{3.53}  & 3.55  & 3.59  & 3.57 \\
42.5 M  & \textbf{3.41}  & 3.44  & 3.46  & 3.45 \\
117.9 M & \textbf{3.25}  & 3.27  & 3.29  & 3.28 \\
1208 M  & \textbf{3.05}  & \texttt{NaN}  & \texttt{NaN}  & 3.08 \\ \bottomrule

\end{tabular}
\end{table}

\textbf{Observations.} We find that in both the training regimes, our proposed algorithms, namely \precond\,\funcSAM and \precondSAMbrief, significantly outperform \SAM  as well as \adamw, at all the model scales in Tables~\ref{tab:fixed-10k} and~\ref{tab:chincilla}. We find that  \SAM performs the worst of the lot, even worse than \adamw while being $2\times$ computationally expensive. Overall,  \precond\,\funcSAM achieves the best results, followed by \precondSAMbrief. %

At the billion-parameter scale, \precondSAMbrief and \SAM show susceptibility to numerical instabilities (yielding \texttt{NaN}s) across training regimes, whereas \textit{\precond\,\funcSAM is significantly more robust}. In the relatively cheaper fixed-length runs where we could carry out additional investigation, we find that warming up the perturbation radius can mitigate these issues for \precondSAMbrief and \SAM, yielding an evaluation loss of 3.07 and 3.08, respectively. 
However, they are still outperformed by our best method, \precond\,\funcSAM, which obtains a lower evaluation loss of $3.05$. and does not need a perturbation warm-up in either training regime. %
\looseness=-1

\begin{table}[ht]
\caption{Evaluation loss comparison of different methods in \textit{Chinchilla like training~\citep{everett2024scaling}}. Lower is better.\looseness=-1}
\renewcommand{\arraystretch}{1.2} %
\setlength{\tabcolsep}{1pt} %
\centering
\label{tab:chincilla}
\begin{tabular}{@{}lcccc@{}}
\toprule
\textsc{Size} &\begin{tabular}[c]{@{}c@{}}\precond\\   \funcSAMbrief\end{tabular} & \begin{tabular}[c]{@{}c@{}}\precond \\ \SAM\end{tabular} & \SAM  & \adamw  \\ \midrule
23.9 M  & \textbf{3.63}  & 3.66  & 3.71  & 3.69 \\
42.5 M  & \textbf{3.41}  & 3.43  & 3.46  & 3.45 \\
117.9 M & \textbf{3.10}  & 3.11  & 3.13  & 3.12 \\
1208 M  & \textbf{2.61}  & \texttt{NaN}   & \texttt{NaN}   & 2.63 \\ \bottomrule
\end{tabular}
\end{table}

Also, it should be noted that in the case of larger models, we tune the perturbation radius $\rho$ at a much coarser level to reduce the associated experimental costs. But despite this likely sub-optimal tuning of the $\rho$ for larger models, we nevertheless observe consistent gains in performance in both fixed-length and Chinchilla like training regimes.

\textbf{Discussion.} The above-mentioned results demonstrate that the issue with \SAM in NLP can be successfully resolved through \precond\,\funcSAM. 
The significant improvements in validation performance above are \textit{especially intriguing }if we bear in mind that these are obtained (a) even when using a \textit{clean corpus such as C4} and (b) training in an \textit{online fashion where no batch is seen more than once} due to massive size of the C4 corpus (even when training for longer duration like in the Chinchilla setup). In industrial training of LLMs, these are idealized conditions which may not always hold and so we can expect further improvements in generalization when training is carried out on noisy corpora and where some parts of it are seen multiple times. \looseness=-1

All in all, these results confirm the benefits imparted by \funcSAM and \precondSAM over \SAM, and these improvements in generalization over \adamw \textit{restore the promise of sharpness regularization.}\looseness=-1%

\begin{table*}[ht!]
    \centering
     \caption{Comparison of different methods based on Hessian ${\HL}$ and GGN ${\HG}$ maximum eigenvalue and trace for the $23.9$M model trained as per Chinchilla like training setup. Lower is better for all metrics. The best entry is in \textbf{bold}, the second best is \underline{underlined}.\looseness=-1}%
    \renewcommand{\arraystretch}{1.2} %
\setlength{\tabcolsep}{3.5pt} %

    \begin{tabular}{lccp{1.5cm}c}
    \toprule
    \textsc{Method} & \textsc{Eval Loss} & $\,\,\lambda_{\text{max}}({\HL})$ & $\,\,\mathrm{tr}({\HL})$ & $\mathrm{tr}({\HG})$ \\
    \midrule
    \adamw & 3.69 & 10.61 & 4897.52 & 4745.03 \\[2mm]
    \SAM & 3.71 & \textbf{2.71} & 3324.58 & 3231.46 \\[1mm]\midrule
    \makecell[l]{\precond \\SAM} & \underline{3.66} & \underline{5.62} & \underline{3182.78} & \underline{3097.87} \\[3mm]
    \makecell[l]{\precond\\\funcSAMbrief} & \multirow[c]{1}{*}{\textbf{3.63}} & 6.20 & \textbf{2687.12} & \textbf{2503.86} \\
    \bottomrule
\end{tabular}

    \label{tab:hessian_comparison_23.9M}
\end{table*}

\subsection{Ablation Studies}
\begin{figure}[ht!]
\vskip 0.1in
\centering
\begin{center}
\centerline{\includegraphics[trim=5 10 0 0,clip,width=\columnwidth]{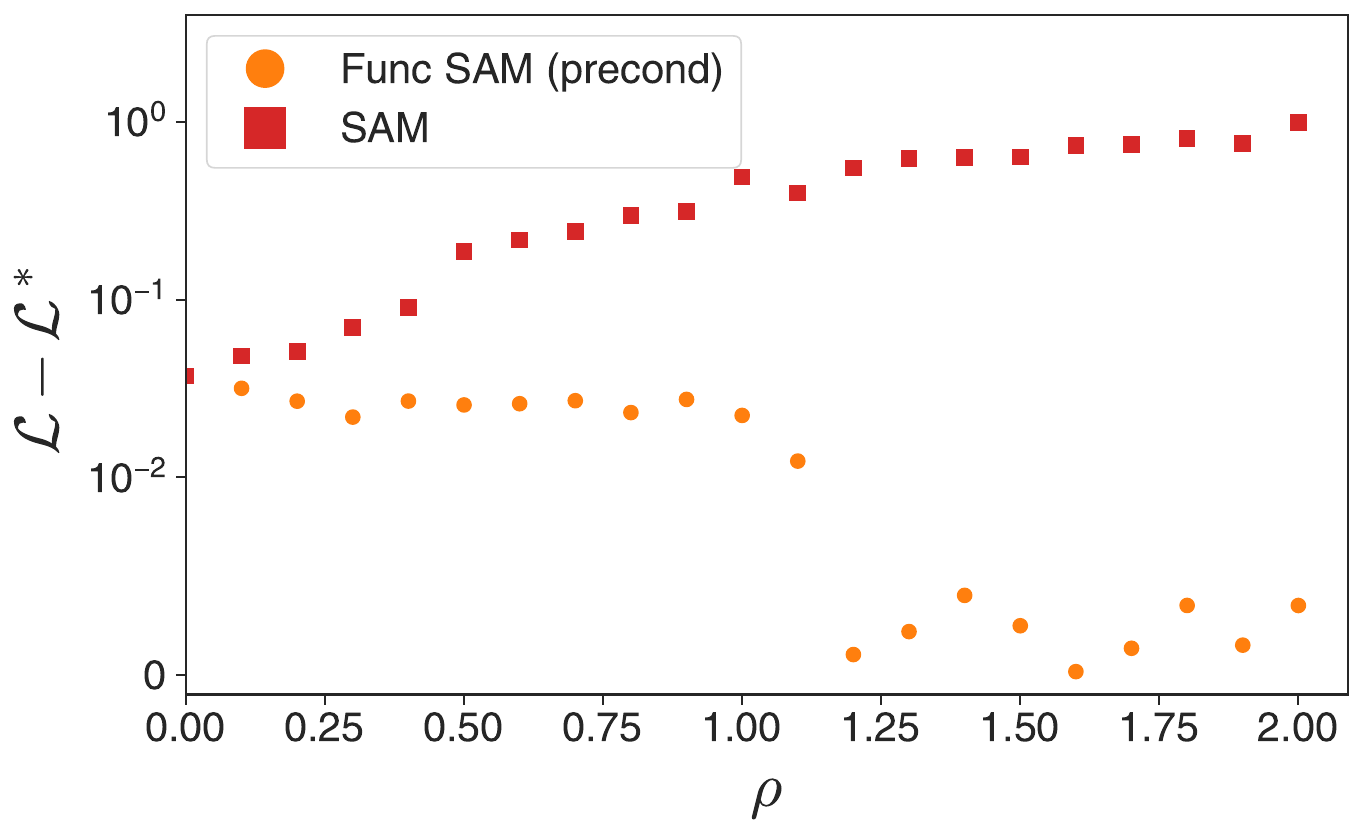}}
\caption{\textit{Effect of increasing perturbation strengths for \textcolor{samColor}{\textsc{SAM}} and \fcolor{\textsc{precond Functional-SAM}} at equal compute}. We see that \SAM (red squares) does worse than baseline at non-zero $\rho$, while {\textsc{precond Functional-SAM}} (orange circles), with the same compute costs, shows improvements (Nanodo trained on C4, $23.9$M parameters, $10$K steps). Loss is measured relative to best validation metric for \precond\,\funcSAM for illustrative purposes.\looseness=-1}
\label{fig:sam_fn_sam_rho}
\end{center}
\vskip -0.2in 
\end{figure}

\textbf{Perturbation strengths.} In the above-mentioned results, the best values for \SAM typically occur around perturbation radius $\rho=0.1$, which, in our experiments, happens to be the smallest nonzero perturbation radius considered. However, if we employ larger perturbation radii, the performance of \SAM rapidly deteriorates, as shown in Figure~\ref{fig:sam_fn_sam_rho}, which moreover suggests that the optimal value of perturbation size $\rho$ is $0$ --- i.e., \textit{not using \SAM at all}. In contrast, the optimal value of perturbation for \precond\,\funcSAM tends to be much larger.
This explains why we needed to modify the relative magnitudes of the sharpness contributions using \precond\,\funcSAM ---
the logit term degrades performance at even small $\rad$, overwhelming the potential gains from the functional path to sharpness minimization.
\looseness=-1

\textbf{Non-linearity Choice.} While all the above experimental results utilize GeLU non-linearity, we also carry out experiments with ReLU as the choice of non-linearity. These results can be found in Table~\ref{tab:non-linearities} of the Appendix, but the key observation is that similar level of improvement is observed in the evaluation loss when using \precond\,\funcSAM for ReLU based architectures as well.

\subsection{Sharpness of Final Solutions}
Here, we confirm that the generalization benefits imparted by \precond\,\funcSAM are brought about convergence to a solution with lower curvature, as shown in the Table~\ref{tab:hessian_comparison_23.9M} for the $23.9$M model. Similar results can also be found on other model sizes, and Table~\ref{tab:hessian_comparison_117.9M} of the Appendix shows them for the $117.9$M model.\looseness=-1

We notice that all sharpness minimization methods yield lower value of the maximum eigenvalue and the trace, as compared to \adamw. Interestingly, we also that \SAM results in the lowest Hessian maximum eigenvalue, even though it performs even worse than \adamw when evaluating on the validation set. This further highlights how \SAM, by default, in language modeling tasks is set up to minimize sharpness spuriously. In contrast, we see that for both of our proposed methods, the improved generalized performance comes hand-in-hand with better landscape properties of their solutions. \looseness=-1

\section{Related work}

\textbf{Improved variants of \SAM in Vision.} An extensive line of work has attempted to propose better definitions of sharpness --- particularly those which are less sensitive to
details of parameterization
\citep{kwon2021asam,tahmasebi2024universalclasssharpnessawareminimization,li2024enhancing}. Some of these methods 
have shown small improvements on vision tasks. \textit{We believe our decomposition approach is orthogonal to this line of research.} Therefore, the obtained \funcSAM algorithm is substantially different from  prior work.

\textbf{Studies exploring preconditioning for \SAM.} Other work has also suggested that the perturbation step in \SAM should be taken in an alternative geometry. Our approach to preconditioning is most
similar to Fisher \SAM~\citep{kim2022fisher}, and the concurrent work of~\citet{zhang2025preconditioned} which describes a more
general preconditioning scheme for \SAM. Our key insight is that it is useful to take the \SAM perturbation in the \emph{exact same geometry}
used by the optimizer, which can be accomplished for negligible cost in the case of Adam and its variants.
Furthermore, our work is \textit{primarily driven by the problem of making \SAM work in language modeling}, which is far from the focus of these other works.\looseness=-1

\textbf{Role of the indefinite Hessian term.} At a more conceptual level, our study aligns with recent works~\citep{singh2021analytic,dauphin2024neglected} which underscore paying more importance to the functional Hessian, the understudied indefinite term of the Hessian in the Gauss-Newton decomposition, as opposed to focusing solely on the positive semi-definite GGN term as suggested by prior studies~\citep{sagun2018empiricalanalysishessianoverparametrized,papyan2019spectrumdeepnethessiansscale,jacot2020asymptoticspectrumhessiandnn}. The decomposition of the sharpness gradient into logit and functional modes, along with the demonstrated significance of \funcSAM in NLP tasks (which is closely tied to the functional Hessian),\textit{ highlights the risks of over-reliance on the GGN and the corresponding outlier spectrum of the Hessian} --- 
especially in the context of \emph{regularization} of sharpness.

\textbf{\SAM in NLP.} Prior works building on \SAM in NLP have been restricted to the fine-tuning setting~\citep{bahri2022sharpnessawareminimizationimproveslanguage}, domain transfer~\citep{sherborne2024trambridgingtrustregions} or on small-scale machine translation setups~\citep{li2024enhancing}.
To the best of our knowledge, \SAM has hitherto not been successfully applied to language modeling, particularly in any scaling
setting.

\section{Discussion and Future Work}\label{sec:disc}
There are several important aspects that we would like to elaborate on, which could spark interesting future work:\looseness=-1

\textbf{Interpolating smoothly between paths to sharpness minimization.} In this work, we focused on \funcSAM due
to our diagnosis of issues in language modeling. However, one can imagine that there might be other domains where \funcSAM gives
spurious minimization, and the alternative ``\logitSAM'' may need to be emphasized. Our implementation of \funcSAM can be extended
to a more continuous ``\angleSAM'' which can smoothly interpolate between the extremes, which we discuss in detail in Appendix~\ref{app:angle-sam}. Though we did not
find \angleSAM necessary in the language modeling setting, there might be other model-dataset-optimizer triples where it would prove beneficial.

\textbf{The Perturbation Scope.} One of the key takeaways from this work is how scoping the perturbation to the level of the function Jacobian suffices to enable the benefits of \SAM in NLP tasks.
This scoping can be generalized to \emph{any}
set of paths or branches in the computational graph. For example, the output of network with residual connections can be decomposed into multiple streams like $f(\x) = f_1(\x) + f_2(\x) + \cdots + f_m(\x)$ and
practitioners can choose to either have the perturbation go through all or some of them. This perspective opens the door to many more creative levels of perturbation scoping and new regularization techniques.

\textbf{Optimal Perturbation Transfer.} At the largest scales, extensive tuning of the perturbation radius is unfeasible; \funcSAM and its
variants will benefit from improved parameterizations that optimally transfer the perturbation radius across model scales.  The recently explored layerwise perturbation scaling regime for \SAM in vision tasks~\citep{haas2024boldsymbolmumathbfp2effectivesharpnessaware} may be promising for language too.
\looseness=-1

\textbf{Utility for Downstream tasks and deployment.}
An added advantage of training with sharpness minimization methods is that the resulting flatter solutions can adapt more gracefully to the post-training pipeline, like downstream fine-tuning or model compression. These benefits have been demonstrated in prior work such as~\citep{liu2023same}, where a lower Hessian trace at the solution has been shown to correlate better with performance on downstream tasks, and we expect similar benefits to also hold with \funcSAM and its variants, given the obtained flatter solutions (see Tables~\ref{tab:hessian_comparison_23.9M},~\ref{tab:hessian_comparison_117.9M}). Likewise prior work~\citep{Na_2022} has also advocated the use of such methods when subsequent model compression is intended, and we also present a very simple demonstration with one-shot pruning in Figure~\ref{fig:pruning} where we see that \funcSAM shows a more graceful degradation as opposed to \adamw with increasing sparsity. We leave a detailed study  to future work.\looseness=-1 %

\textbf{Efficiency.} A common drawback of sharpness minimization methods is that they require twice the gradient computation per step, and hence are twice as expensive  as compared to other optimizers like \adamw. In this paper, our singular focus  has been to address the ineffectiveness of \SAM in language modeling, which we have been able to carry out successfully through the proposed \funcSAM and its preconditioning variants. However, these algorithms still share the same $2\times$ computational burden like \SAM itself, and thus additional work is required in the future to make it suitable for deployment.
Thankfully, since \funcSAM is structured similarly to \SAM, most advancements in efficient implementations of \SAM should be
useful for \funcSAM as well. Given the plethora of recent work in this area for vision 
~\citep{du2022sharpness,liu2022towards,becker2024momentumsamsharpnessawareminimization,xie2024sampasharpnessawareminimizationparallelized}, we are
optimistic about the development of efficient \funcSAM variants for language modeling in the near future.

\section{Conclusion}
Our work thoroughly highlights the ineffectiveness of \SAM for language modeling and uncovers the underlying reasons behind such an 
occurrence. We show that this arises because sharpness reduction in NLP settings is prone to logit manipulation, and hence spurious sharpness
minimization, --- rather than being driven by promoting the simplicity of the network function. Based on this insight, 
we propose addressing this issue via \funcSAM and preconditioning. We demonstrate that both these simple but  principled modifications to \SAM restore its generalization properties across multiple model scales. 

More broadly, we believe that our novel foray into  functional and logit modes of sharpness reduction will reinvigorate the existing research into \SAM, and pave the way for advanced curvature regularization techniques. 
Lastly, we are excited about the nuanced characterization of sharpness introduced here and hope that it advances our fundamental understanding of sharpness, its dynamics, and the broader nature of loss landscapes.

\clearpage

%% file: code.tex
\lstset{style=mystyle}
\begin{lstlisting}[language=Python, caption={Illustration of how to get the gradients in the two methods. Functional SAM differs from the SAM implementation only in the last couple lines, where the effect of perturbation is made to reside only in the function Jacobian part.}]
from jax import grad, vjp
from jax.tree_util import tree_map
from utils import normalize_grad

def sam_gradients(params, loss_fn, rho):

    # compute the usual loss gradient
    dL_dtheta = grad(loss_fn)(params)
    
    # normalize the gradients
    dL_dtheta = normalize_grad(dL_dtheta)
    
    # perturb the parameters
    perturbed_params = tree_map(lambda a, b: a + rho * b, params, dL_dtheta)
    
    # compute the gradient as by SAM
    sam_grad = grad(loss_fn)(perturbed_params)
    
    return sam_grad

def functional_sam_gradients(params, loss_fn, network_fn, rho):
    
    # compute the usual loss gradient, but also extract dL_dlogits 
    (dL_dlogits), dL_dtheta = grad(loss_fn, hax_aux=True)(params)
    
    # normalize the gradients
    dL_dtheta = normalize_grad(dL_dtheta)
    
    # perturb the parameters
    perturbed_params = tree_map(lambda a, b: a + rho * b, params, dL_dtheta)
    
    # set up the VJP at the perturbed parameters
    _, dF_dtheta_fn = vjp(lambda theta: network_fn(theta), perturbed_params)
    
    # do the VJP with the (unperturbed) dL_dlogits
    functional_sam_grad = dF_dtheta_fn(dL_dlogits)[0]
    
    return functional_sam_grad

\end{lstlisting}

%% file: googledeepmind-test.bbl
\begin{thebibliography}{62}
\providecommand{\natexlab}[1]{#1}
\providecommand{\url}[1]{\texttt{#1}}
\expandafter\ifx\csname urlstyle\endcsname\relax
  \providecommand{\doi}[1]{doi: #1}\else
  \providecommand{\doi}{doi: \begingroup \urlstyle{rm}\Url}\fi

\bibitem[Andriushchenko and
  Flammarion(2022)]{andriushchenko2022understandingsharpnessawareminimization}
M.~Andriushchenko and N.~Flammarion.
\newblock Towards understanding sharpness-aware minimization, 2022.
\newblock URL \url{https://arxiv.org/abs/2206.06232}.

\bibitem[Bahri et~al.(2022)Bahri, Mobahi, and
  Tay]{bahri2022sharpnessawareminimizationimproveslanguage}
D.~Bahri, H.~Mobahi, and Y.~Tay.
\newblock Sharpness-aware minimization improves language model generalization,
  2022.
\newblock URL \url{https://arxiv.org/abs/2110.08529}.

\bibitem[Becker et~al.(2024)Becker, Altrock, and
  Risse]{becker2024momentumsamsharpnessawareminimization}
M.~Becker, F.~Altrock, and B.~Risse.
\newblock Momentum-sam: Sharpness aware minimization without computational
  overhead, 2024.
\newblock URL \url{https://arxiv.org/abs/2401.12033}.

\bibitem[Bradbury et~al.(2018)Bradbury, Frostig, Hawkins, Johnson, Leary,
  Maclaurin, Necula, Paszke, Vander{P}las, Wanderman-{M}ilne, and
  Zhang]{jax2018github}
J.~Bradbury, R.~Frostig, P.~Hawkins, M.~J. Johnson, C.~Leary, D.~Maclaurin,
  G.~Necula, A.~Paszke, J.~Vander{P}las, S.~Wanderman-{M}ilne, and Q.~Zhang.
\newblock {JAX}: composable transformations of {P}ython+{N}um{P}y programs,
  2018.
\newblock URL \url{http://github.com/jax-ml/jax}.

\bibitem[Brown et~al.(2020)Brown, Mann, Ryder, Subbiah, Kaplan, Dhariwal,
  Neelakantan, Shyam, Sastry, Askell, Agarwal, Herbert-Voss, Krueger, Henighan,
  Child, Ramesh, Ziegler, Wu, Winter, Hesse, Chen, Sigler, Litwin, Gray, Chess,
  Clark, Berner, McCandlish, Radford, Sutskever, and
  Amodei]{brown2020languagemodelsfewshotlearners}
T.~B. Brown, B.~Mann, N.~Ryder, M.~Subbiah, J.~Kaplan, P.~Dhariwal,
  A.~Neelakantan, P.~Shyam, G.~Sastry, A.~Askell, S.~Agarwal, A.~Herbert-Voss,
  G.~Krueger, T.~Henighan, R.~Child, A.~Ramesh, D.~M. Ziegler, J.~Wu,
  C.~Winter, C.~Hesse, M.~Chen, E.~Sigler, M.~Litwin, S.~Gray, B.~Chess,
  J.~Clark, C.~Berner, S.~McCandlish, A.~Radford, I.~Sutskever, and D.~Amodei.
\newblock Language models are few-shot learners, 2020.
\newblock URL \url{https://arxiv.org/abs/2005.14165}.

\bibitem[Chaudhari et~al.(2017)Chaudhari, Choromanska, Soatto, LeCun, Baldassi,
  Borgs, Chayes, Sagun, and
  Zecchina]{chaudhari2017entropysgdbiasinggradientdescent}
P.~Chaudhari, A.~Choromanska, S.~Soatto, Y.~LeCun, C.~Baldassi, C.~Borgs,
  J.~Chayes, L.~Sagun, and R.~Zecchina.
\newblock Entropy-sgd: Biasing gradient descent into wide valleys, 2017.
\newblock URL \url{https://arxiv.org/abs/1611.01838}.

\bibitem[Ciregan et~al.(2012)Ciregan, Meier, and Schmidhuber]{ciregan2012multi}
D.~Ciregan, U.~Meier, and J.~Schmidhuber.
\newblock Multi-column deep neural networks for image classification.
\newblock In \emph{2012 IEEE conference on computer vision and pattern
  recognition}, pages 3642--3649. IEEE, 2012.

\bibitem[Dauphin et~al.(2024)Dauphin, Agarwala, and
  Mobahi]{dauphin2024neglected}
Y.~N. Dauphin, A.~Agarwala, and H.~Mobahi.
\newblock Neglected hessian component explains mysteries in sharpness
  regularization, 2024.

\bibitem[Dosovitskiy et~al.(2021)Dosovitskiy, Beyer, Kolesnikov, Weissenborn,
  Zhai, Unterthiner, Dehghani, Minderer, Heigold, Gelly, Uszkoreit, and
  Houlsby]{dosovitskiy2021imageworth16x16words}
A.~Dosovitskiy, L.~Beyer, A.~Kolesnikov, D.~Weissenborn, X.~Zhai,
  T.~Unterthiner, M.~Dehghani, M.~Minderer, G.~Heigold, S.~Gelly, J.~Uszkoreit,
  and N.~Houlsby.
\newblock An image is worth 16x16 words: Transformers for image recognition at
  scale, 2021.
\newblock URL \url{https://arxiv.org/abs/2010.11929}.

\bibitem[Du et~al.(2022)Du, Zhou, Feng, Tan, and Zhou]{du2022sharpness}
J.~Du, D.~Zhou, J.~Feng, V.~Y. Tan, and J.~T. Zhou.
\newblock Sharpness-aware training for free.
\newblock \emph{arXiv preprint arXiv:2205.14083}, 2022.

\bibitem[Everett et~al.(2024)Everett, Xiao, Wortsman, Alemi, Novak, Liu, Gur,
  Sohl-Dickstein, Kaelbling, Lee, et~al.]{everett2024scaling}
K.~Everett, L.~Xiao, M.~Wortsman, A.~A. Alemi, R.~Novak, P.~J. Liu, I.~Gur,
  J.~Sohl-Dickstein, L.~P. Kaelbling, J.~Lee, et~al.
\newblock Scaling exponents across parameterizations and optimizers.
\newblock \emph{arXiv preprint arXiv:2407.05872}, 2024.

\bibitem[Foret et~al.(2020)Foret, Kleiner, Mobahi, and
  Neyshabur]{foret2020sharpness}
P.~Foret, A.~Kleiner, H.~Mobahi, and B.~Neyshabur.
\newblock Sharpness-aware minimization for efficiently improving
  generalization.
\newblock \emph{arXiv preprint arXiv:2010.01412}, 2020.

\bibitem[Gur-Ari et~al.(2018)Gur-Ari, Roberts, and Dyer]{gur2018gradient}
G.~Gur-Ari, D.~A. Roberts, and E.~Dyer.
\newblock Gradient descent happens in a tiny subspace.
\newblock \emph{arXiv preprint arXiv:1812.04754}, 2018.

\bibitem[Haas et~al.(2024)Haas, Xu, Cevher, and
  Vankadara]{haas2024boldsymbolmumathbfp2effectivesharpnessaware}
M.~Haas, J.~Xu, V.~Cevher, and L.~C. Vankadara.
\newblock $\boldsymbol{\mu}\mathbf{P^2}$: Effective sharpness aware
  minimization requires layerwise perturbation scaling, 2024.
\newblock URL \url{https://arxiv.org/abs/2411.00075}.

\bibitem[Heek et~al.(2024)Heek, Levskaya, Oliver, Ritter, Rondepierre, Steiner,
  and van {Z}ee]{flax2020github}
J.~Heek, A.~Levskaya, A.~Oliver, M.~Ritter, B.~Rondepierre, A.~Steiner, and
  M.~van {Z}ee.
\newblock {F}lax: A neural network library and ecosystem for {JAX}, 2024.
\newblock URL \url{http://github.com/google/flax}.

\bibitem[Hinton and Van~Camp(1993)]{hinton1993keeping}
G.~E. Hinton and D.~Van~Camp.
\newblock Keeping the neural networks simple by minimizing the description
  length of the weights.
\newblock In \emph{Proceedings of the sixth annual conference on Computational
  learning theory}, pages 5--13, 1993.

\bibitem[Hochreiter and Schmidhuber(1997)]{Hochreiter1997FlatM}
S.~Hochreiter and J.~Schmidhuber.
\newblock Flat minima.
\newblock \emph{Neural Computation}, 9:\penalty0 1--42, 1997.
\newblock URL \url{https://api.semanticscholar.org/CorpusID:733161}.

\bibitem[Hoffmann et~al.(2022)Hoffmann, Borgeaud, Mensch, Buchatskaya, Cai,
  Rutherford, de~Las~Casas, Hendricks, Welbl, Clark, Hennigan, Noland,
  Millican, van~den Driessche, Damoc, Guy, Osindero, Simonyan, Elsen, Rae,
  Vinyals, and Sifre]{hoffmann2022trainingcomputeoptimallargelanguage}
J.~Hoffmann, S.~Borgeaud, A.~Mensch, E.~Buchatskaya, T.~Cai, E.~Rutherford,
  D.~de~Las~Casas, L.~A. Hendricks, J.~Welbl, A.~Clark, T.~Hennigan, E.~Noland,
  K.~Millican, G.~van~den Driessche, B.~Damoc, A.~Guy, S.~Osindero,
  K.~Simonyan, E.~Elsen, J.~W. Rae, O.~Vinyals, and L.~Sifre.
\newblock Training compute-optimal large language models, 2022.
\newblock URL \url{https://arxiv.org/abs/2203.15556}.

\bibitem[Jacot et~al.(2020)Jacot, Gabriel, and
  Hongler]{jacot2020asymptoticspectrumhessiandnn}
A.~Jacot, F.~Gabriel, and C.~Hongler.
\newblock The asymptotic spectrum of the hessian of dnn throughout training,
  2020.
\newblock URL \url{https://arxiv.org/abs/1910.02875}.

\bibitem[Jiang et~al.(2024)Jiang, Malik, and Li]{jiang2024does}
K.~Jiang, D.~Malik, and Y.~Li.
\newblock How does adaptive optimization impact local neural network geometry?
\newblock \emph{Advances in Neural Information Processing Systems}, 36, 2024.

\bibitem[Jiang et~al.(2019)Jiang, Neyshabur, Mobahi, Krishnan, and
  Bengio]{jiang2019fantasticgeneralizationmeasures}
Y.~Jiang, B.~Neyshabur, H.~Mobahi, D.~Krishnan, and S.~Bengio.
\newblock Fantastic generalization measures and where to find them, 2019.
\newblock URL \url{https://arxiv.org/abs/1912.02178}.

\bibitem[Jiang et~al.(2020)Jiang, Foret, Yak, Roy, Mobahi, Dziugaite, Bengio,
  Gunasekar, Guyon, and Neyshabur]{jiang2020neurips2020competitionpredicting}
Y.~Jiang, P.~Foret, S.~Yak, D.~M. Roy, H.~Mobahi, G.~K. Dziugaite, S.~Bengio,
  S.~Gunasekar, I.~Guyon, and B.~Neyshabur.
\newblock Neurips 2020 competition: Predicting generalization in deep learning,
  2020.
\newblock URL \url{https://arxiv.org/abs/2012.07976}.

\bibitem[Kaplan et~al.(2020)Kaplan, McCandlish, Henighan, Brown, Chess, Child,
  Gray, Radford, Wu, and Amodei]{kaplan2020scalinglawsneurallanguage}
J.~Kaplan, S.~McCandlish, T.~Henighan, T.~B. Brown, B.~Chess, R.~Child,
  S.~Gray, A.~Radford, J.~Wu, and D.~Amodei.
\newblock Scaling laws for neural language models, 2020.
\newblock URL \url{https://arxiv.org/abs/2001.08361}.

\bibitem[Keskar et~al.(2017)Keskar, Mudigere, Nocedal, Smelyanskiy, and
  Tang]{keskar2017largebatchtrainingdeeplearning}
N.~S. Keskar, D.~Mudigere, J.~Nocedal, M.~Smelyanskiy, and P.~T.~P. Tang.
\newblock On large-batch training for deep learning: Generalization gap and
  sharp minima, 2017.
\newblock URL \url{https://arxiv.org/abs/1609.04836}.

\bibitem[Kim et~al.(2022)Kim, Li, Hu, and Hospedales]{kim2022fisher}
M.~Kim, D.~Li, S.~X. Hu, and T.~Hospedales.
\newblock Fisher sam: Information geometry and sharpness aware minimisation.
\newblock In \emph{International Conference on Machine Learning}, pages
  11148--11161. PMLR, 2022.

\bibitem[Kingma and Ba(2017)]{kingma2017adammethodstochasticoptimization}
D.~P. Kingma and J.~Ba.
\newblock Adam: A method for stochastic optimization, 2017.
\newblock URL \url{https://arxiv.org/abs/1412.6980}.

\bibitem[Krizhevsky et~al.(2012)Krizhevsky, Sutskever, and
  Hinton]{krizhevsky2012imagenet}
A.~Krizhevsky, I.~Sutskever, and G.~E. Hinton.
\newblock Imagenet classification with deep convolutional neural networks.
\newblock \emph{Advances in neural information processing systems}, 25, 2012.

\bibitem[Krogh and Hertz(1991)]{krogh1991simple}
A.~Krogh and J.~Hertz.
\newblock A simple weight decay can improve generalization.
\newblock \emph{Advances in neural information processing systems}, 4, 1991.

\bibitem[Kwon et~al.(2021)Kwon, Kim, Park, and Choi]{kwon2021asam}
J.~Kwon, J.~Kim, H.~Park, and I.~K. Choi.
\newblock Asam: Adaptive sharpness-aware minimization for scale-invariant
  learning of deep neural networks.
\newblock In \emph{International Conference on Machine Learning}, pages
  5905--5914. PMLR, 2021.

\bibitem[Li and Giannakis(2024)]{li2024enhancing}
B.~Li and G.~Giannakis.
\newblock Enhancing sharpness-aware optimization through variance suppression.
\newblock \emph{Advances in Neural Information Processing Systems}, 36, 2024.

\bibitem[Liu et~al.(2023)Liu, Xie, Li, and Ma]{liu2023same}
H.~Liu, S.~M. Xie, Z.~Li, and T.~Ma.
\newblock Same pre-training loss, better downstream: Implicit bias matters for
  language models, 2023.
\newblock URL \url{https://openreview.net/forum?id=F5uYcwABMu}.

\bibitem[Liu et~al.(2020)Liu, Liu, Gao, Chen, and Han]{liu2020understanding}
L.~Liu, X.~Liu, J.~Gao, W.~Chen, and J.~Han.
\newblock Understanding the difficulty of training transformers.
\newblock \emph{arXiv preprint arXiv:2004.08249}, 2020.

\bibitem[Liu et~al.(2024)Liu, Novak, Lee, Wortsman, Xiao, Everett, Alemi,
  Kurzeja, Marcenac, Gur, Kornblith, Xu, Elsayed, Fischer, Pennington, Adlam,
  and Dickstein]{nanodo}
P.~J. Liu, R.~Novak, J.~Lee, M.~Wortsman, L.~Xiao, K.~Everett, A.~A. Alemi,
  M.~Kurzeja, P.~Marcenac, I.~Gur, S.~Kornblith, K.~Xu, G.~Elsayed, I.~Fischer,
  J.~Pennington, B.~Adlam, and J.-S. Dickstein.
\newblock Nanodo: A minimal transformer decoder-only language model
  implementation in {JAX}., 2024.
\newblock URL \url{http://github.com/google-deepmind/nanodo}.

\bibitem[Liu et~al.(2022)Liu, Mai, Chen, Hsieh, and You]{liu2022towards}
Y.~Liu, S.~Mai, X.~Chen, C.-J. Hsieh, and Y.~You.
\newblock Towards efficient and scalable sharpness-aware minimization.
\newblock In \emph{Proceedings of the IEEE/CVF Conference on Computer Vision
  and Pattern Recognition}, pages 12360--12370, 2022.

\bibitem[Loshchilov and
  Hutter(2019)]{loshchilov2019decoupledweightdecayregularization}
I.~Loshchilov and F.~Hutter.
\newblock Decoupled weight decay regularization, 2019.
\newblock URL \url{https://arxiv.org/abs/1711.05101}.

\bibitem[Na et~al.(2022)Na, Mehta, and Strubell]{Na_2022}
C.~Na, S.~V. Mehta, and E.~Strubell.
\newblock Train flat, then compress: Sharpness-aware minimization learns more
  compressible models.
\newblock In \emph{Findings of the Association for Computational Linguistics:
  EMNLP 2022}, page 4909–4936. Association for Computational Linguistics,
  2022.
\newblock \doi{10.18653/v1/2022.findings-emnlp.361}.
\newblock URL \url{http://dx.doi.org/10.18653/v1/2022.findings-emnlp.361}.

\bibitem[Noci et~al.(2022)Noci, Anagnostidis, Biggio, Orvieto, Singh, and
  Lucchi]{noci2022signal}
L.~Noci, S.~Anagnostidis, L.~Biggio, A.~Orvieto, S.~P. Singh, and A.~Lucchi.
\newblock Signal propagation in transformers: Theoretical perspectives and the
  role of rank collapse.
\newblock \emph{Advances in Neural Information Processing Systems},
  35:\penalty0 27198--27211, 2022.

\bibitem[Ormaniec et~al.(2024)Ormaniec, Dangel, and
  Singh]{ormaniec2024doesmeantransformerinsights}
W.~Ormaniec, F.~Dangel, and S.~P. Singh.
\newblock What does it mean to be a transformer? insights from a theoretical
  hessian analysis, 2024.
\newblock URL \url{https://arxiv.org/abs/2410.10986}.

\bibitem[Pan and Li(2023)]{pan2023toward}
Y.~Pan and Y.~Li.
\newblock Toward understanding why adam converges faster than sgd for
  transformers.
\newblock \emph{arXiv preprint arXiv:2306.00204}, 2023.

\bibitem[Papyan(2019)]{papyan2019spectrumdeepnethessiansscale}
V.~Papyan.
\newblock The full spectrum of deepnet hessians at scale: Dynamics with sgd
  training and sample size, 2019.
\newblock URL \url{https://arxiv.org/abs/1811.07062}.

\bibitem[Pennington and Bahri(2017)]{pennington17rmt}
J.~Pennington and Y.~Bahri.
\newblock Geometry of neural network loss surfaces via random matrix theory.
\newblock In D.~Precup and Y.~W. Teh, editors, \emph{Proceedings of the 34th
  International Conference on Machine Learning}, volume~70 of \emph{Proceedings
  of Machine Learning Research}, pages 2798--2806. PMLR, 06--11 Aug 2017.
\newblock URL \url{https://proceedings.mlr.press/v70/pennington17a.html}.

\bibitem[Pittorino et~al.(2020)Pittorino, Lucibello, Feinauer, Malatesta,
  Perugini, Baldassi, Negri, Demyanenko, and
  Zecchina]{DBLP:journals/corr/abs-2006-07897}
F.~Pittorino, C.~Lucibello, C.~Feinauer, E.~M. Malatesta, G.~Perugini,
  C.~Baldassi, M.~Negri, E.~Demyanenko, and R.~Zecchina.
\newblock Entropic gradient descent algorithms and wide flat minima.
\newblock \emph{CoRR}, abs/2006.07897, 2020.
\newblock URL \url{https://arxiv.org/abs/2006.07897}.

\bibitem[Radford et~al.(2019)Radford, Wu, Child, Luan, Amodei, Sutskever,
  et~al.]{radford2019language}
A.~Radford, J.~Wu, R.~Child, D.~Luan, D.~Amodei, I.~Sutskever, et~al.
\newblock Language models are unsupervised multitask learners.
\newblock \emph{OpenAI blog}, 1\penalty0 (8):\penalty0 9, 2019.

\bibitem[Raffel et~al.(2020)Raffel, Shazeer, Roberts, Lee, Narang, Matena,
  Zhou, Li, and Liu]{raffel2020exploring}
C.~Raffel, N.~Shazeer, A.~Roberts, K.~Lee, S.~Narang, M.~Matena, Y.~Zhou,
  W.~Li, and P.~J. Liu.
\newblock Exploring the limits of transfer learning with a unified text-to-text
  transformer.
\newblock \emph{Journal of machine learning research}, 21\penalty0
  (140):\penalty0 1--67, 2020.

\bibitem[Ridnik et~al.(2021)Ridnik, Ben-Baruch, Noy, and
  Zelnik-Manor]{ridnik2021imagenet21kpretrainingmasses}
T.~Ridnik, E.~Ben-Baruch, A.~Noy, and L.~Zelnik-Manor.
\newblock Imagenet-21k pretraining for the masses, 2021.
\newblock URL \url{https://arxiv.org/abs/2104.10972}.

\bibitem[Rissanen(1978)]{rissanen1978modeling}
J.~Rissanen.
\newblock Modeling by shortest data description.
\newblock \emph{Automatica}, 14\penalty0 (5):\penalty0 465--471, 1978.

\bibitem[Sagun et~al.(2018)Sagun, Evci, Guney, Dauphin, and
  Bottou]{sagun2018empiricalanalysishessianoverparametrized}
L.~Sagun, U.~Evci, V.~U. Guney, Y.~Dauphin, and L.~Bottou.
\newblock Empirical analysis of the hessian of over-parametrized neural
  networks, 2018.
\newblock URL \url{https://arxiv.org/abs/1706.04454}.

\bibitem[Schraudolph(2002)]{Schraudolph2002FastCM}
N.~N. Schraudolph.
\newblock Fast curvature matrix-vector products for second-order gradient
  descent.
\newblock \emph{Neural Computation}, 14:\penalty0 1723--1738, 2002.

\bibitem[Sherborne et~al.(2024)Sherborne, Saphra, Dasigi, and
  Peng]{sherborne2024trambridgingtrustregions}
T.~Sherborne, N.~Saphra, P.~Dasigi, and H.~Peng.
\newblock Tram: Bridging trust regions and sharpness aware minimization, 2024.
\newblock URL \url{https://arxiv.org/abs/2310.03646}.

\bibitem[Singh et~al.(2021)Singh, Bachmann, and Hofmann]{singh2021analytic}
S.~P. Singh, G.~Bachmann, and T.~Hofmann.
\newblock Analytic insights into structure and rank of neural network hessian
  maps.
\newblock In A.~Beygelzimer, Y.~Dauphin, P.~Liang, and J.~W. Vaughan, editors,
  \emph{Advances in Neural Information Processing Systems}, 2021.
\newblock URL \url{https://openreview.net/forum?id=otDgw7LM7Nn}.

\bibitem[Singh et~al.(2023)Singh, Hofmann, and
  Sch\"{o}lkopf]{10.5555/3618408.3619732}
S.~P. Singh, T.~Hofmann, and B.~Sch\"{o}lkopf.
\newblock The hessian perspective into the nature of convolutional neural
  networks.
\newblock In \emph{Proceedings of the 40th International Conference on Machine
  Learning}, ICML'23. JMLR.org, 2023.

\bibitem[Srivastava et~al.(2014)Srivastava, Hinton, Krizhevsky, Sutskever, and
  Salakhutdinov]{srivastava2014dropout}
N.~Srivastava, G.~Hinton, A.~Krizhevsky, I.~Sutskever, and R.~Salakhutdinov.
\newblock Dropout: a simple way to prevent neural networks from overfitting.
\newblock \emph{The journal of machine learning research}, 15\penalty0
  (1):\penalty0 1929--1958, 2014.

\bibitem[Sun et~al.(2017)Sun, Shrivastava, Singh, and
  Gupta]{sun2017revisitingunreasonableeffectivenessdata}
C.~Sun, A.~Shrivastava, S.~Singh, and A.~Gupta.
\newblock Revisiting unreasonable effectiveness of data in deep learning era,
  2017.
\newblock URL \url{https://arxiv.org/abs/1707.02968}.

\bibitem[Tahmasebi et~al.(2024)Tahmasebi, Soleymani, Bahri, Jegelka, and
  Jaillet]{tahmasebi2024universalclasssharpnessawareminimization}
B.~Tahmasebi, A.~Soleymani, D.~Bahri, S.~Jegelka, and P.~Jaillet.
\newblock A universal class of sharpness-aware minimization algorithms, 2024.
\newblock URL \url{https://arxiv.org/abs/2406.03682}.

\bibitem[Vapnik(1991)]{vapnik1991principles}
V.~Vapnik.
\newblock Principles of risk minimization for learning theory.
\newblock \emph{Advances in neural information processing systems}, 4, 1991.

\bibitem[Vaswani(2017)]{vaswani2017attention}
A.~Vaswani.
\newblock Attention is all you need.
\newblock \emph{Advances in Neural Information Processing Systems}, 2017.

\bibitem[Vyas et~al.(2024)Vyas, Morwani, Zhao, Shapira, Brandfonbrener, Janson,
  and Kakade]{vyas2024soap}
N.~Vyas, D.~Morwani, R.~Zhao, I.~Shapira, D.~Brandfonbrener, L.~Janson, and
  S.~Kakade.
\newblock Soap: Improving and stabilizing shampoo using adam.
\newblock \emph{arXiv preprint arXiv:2409.11321}, 2024.

\bibitem[Wu et~al.(2020)Wu, tao Xia, and
  Wang]{wu2020adversarialweightperturbationhelps}
D.~Wu, S.~tao Xia, and Y.~Wang.
\newblock Adversarial weight perturbation helps robust generalization, 2020.
\newblock URL \url{https://arxiv.org/abs/2004.05884}.

\bibitem[Xie et~al.(2024)Xie, Pethick, and
  Cevher]{xie2024sampasharpnessawareminimizationparallelized}
W.~Xie, T.~Pethick, and V.~Cevher.
\newblock Sampa: Sharpness-aware minimization parallelized, 2024.
\newblock URL \url{https://arxiv.org/abs/2410.10683}.

\bibitem[Zhang(2017)]{zhang2017mixup}
H.~Zhang.
\newblock mixup: Beyond empirical risk minimization.
\newblock \emph{arXiv preprint arXiv:1710.09412}, 2017.

\bibitem[Zhang et~al.(2024)Zhang, Chen, Ding, Li, Sun, and
  Luo]{zhang2024transformersneedadamhessian}
Y.~Zhang, C.~Chen, T.~Ding, Z.~Li, R.~Sun, and Z.-Q. Luo.
\newblock Why transformers need adam: A hessian perspective, 2024.
\newblock URL \url{https://arxiv.org/abs/2402.16788}.

\bibitem[Zhang et~al.(2025)Zhang, Li, and Giannakis]{zhang2025preconditioned}
Y.~Zhang, B.~Li, and G.~B. Giannakis.
\newblock Preconditioned sharpness-aware minimization: Unifying analysis and a
  novel learning algorithm.
\newblock \emph{arXiv preprint arXiv:2501.06603}, 2025.

\end{thebibliography}
